\documentclass[]{bytedance_seed}

\usepackage[toc,page,header]{appendix}

\usepackage{minitoc}
\usepackage{makecell}
\usepackage{amssymb}
\usepackage{graphicx}
\usepackage{caption}
\usepackage{subcaption}
\usepackage{bbding}
\usepackage{xcolor}
\usepackage{booktabs}
\usepackage{multirow}
\usepackage{tabularx} 
\usepackage{array}
\usepackage{fancyhdr} 

\newlength\savewidth\newcommand\shline{\noalign{\global\savewidth\arrayrulewidth
  \global\arrayrulewidth 1pt}\hline\noalign{\global\arrayrulewidth\savewidth}}
\newcommand{\tablestyle}[2]{\setlength{\tabcolsep}{#1}\renewcommand{\arraystretch}{#2}\centering\footnotesize}
\renewcommand{\paragraph}[1]{\vspace{1.25mm}\noindent\textbf{#1}}

\newcolumntype{x}[1]{>{\centering\arraybackslash}p{#1pt}}
\newcolumntype{y}[1]{>{\raggedright\arraybackslash}p{#1pt}}
\newcolumntype{z}[1]{>{\raggedleft\arraybackslash}p{#1pt}}

\newcommand{\app}{\raise.17ex\hbox{$\scriptstyle\sim$}}

\definecolor{deemph}{gray}{0.6}

\definecolor{baselinecolor}{gray}{.9}

\usepackage[linesnumbered,ruled,vlined,algo2e]{algorithm2e}
\usepackage{amsmath}
\usepackage{xcolor}

\SetCommentSty{mycommfont}

\title{VideoWorld 2: Learning Transferable Knowledge from Real-world Videos}

\author[1,2,*]{Zhongwei Ren}
\author[2]{Yunchao Wei}
\author[2]{Xiao Yu}
\author[2]{Guixun Luo}
\author[2]{Yao Zhao}
\author[1]{Bingyi Kang}
\author[1]{Jiashi Feng}
\author[1, *, \dagger]{Xiaojie Jin}

\affiliation[1]{ByteDance Seed}
\affiliation[2]{Beijing Jiaotong University}

\contribution[*]{Equal Contribution}
\contribution[\dagger]{Project Lead}

\abstract{
Learning transferable knowledge from unlabeled video data and applying it in new environments is a fundamental capability of intelligent agents. 
This work presents \textbf{VideoWorld~2}, which extends VideoWorld~\cite{ren2025videoworld} and provides the first investigation of learning transferable knowledge for complex, long-horizon tasks directly from raw real-world videos. At its core, VideoWorld~2 introduces a dynamics-enhanced Latent Dynamics Model (dLDM) that decouples action dynamics from visual appearance: a pretrained video diffusion model handles visual appearance modeling, enabling the dLDM to learn latent codes that focus on compact and meaningful task-related dynamics. These latent codes are then modeled autoregressively to learn task policies and support long-horizon reasoning. We evaluate VideoWorld~2 on challenging real-world handcraft making tasks, where prior video generation and latent-dynamics models struggle to operate reliably. Remarkably, VideoWorld~2 achieves up to \textbf{70\% improvement in task success rate} and produces coherent long execution videos. In robotics, we show that VideoWorld~2 can acquire transferable manipulation knowledge from the Open-X dataset, which substantially improves task performance on CALVIN, demonstrating strong cross-domain generalization. This study reveals the potential of learning transferable world knowledge directly from raw videos, with all code, data, and models to be open-sourced for further research.
}

\date{\today}
\correspondence{Xiaojiao Jin, Yunchao Wei}

\checkdata[Project Page]{\href{https://maverickren.github.io/VideoWorld2.github.io/}{\texttt{https://VideoWorld2.github.io/}}}

\begin{document}
\maketitle


\section{Introduction}
Current AI models  primarily learn  knowledge{\footnotemark}   from large-scale text data~\cite{touvron2023llama,gpt,comanici2025gemini,anil2023palm2,chowdhery2023palm,jiang2023mistral,wang2024world,bai2023qwen,yang2024qwen2}. However, text alone cannot fully capture the rich information of the real visual world, including world dynamics, spatial relationships, and underlying physical laws. In contrast, animals in nature can acquire knowledge directly from visual signals, and generalize it to solve tasks across diverse scenarios. For instance, a child can reproduce paper-folding skills demonstrated in a video using different paper materials,  without any language instruction. Given the vast abundance of video content available on the internet, enabling  AI models  to learn \textit{generalizable knowledge}~\cite{huitt2003piaget} from raw video data holds significant promise for scaling  their knowledge acquisition and  is fundamental  to their ability to execute tasks effectively in  both real-world and digital environments.
{\footnotetext{{Following prior work~\cite{ren2025videoworld}, we use ``knowledge'' to refer broadly to the rules, reasoning, and planning abilities required for task completion.}}}


VideoWorld~\cite{ren2025videoworld} is among the first works to explore  learning knowledge from synthetic videos. It investigates the acquisition of rules, as well as reasoning and planning capabilities,  from Go game records and simulated robotics environments. The study demonstrates that  models can learn such knowledge solely from visual signals using an autoregressive video generation paradigm. However, extending this paradigm beyond synthetic   domains remains an open challenge.  Real-world videos exhibit substantial   visual diversity, complex action  dynamics,  and often involve long-horizon, multi-step interactions.  These characteristics prevent   the training approach and model design of VideoWorld  from being  directly applied to realistic settings.
When presented with minute-long, multi-step real-world task videos, VideoWorld  fails to   extract the core task-solving knowledge or generalize it to novel scenarios   through observation alone{\textemdash}even for tasks such as paper folding that are easily mastered by children (See Sec.~\ref{subsec:res_ldm} for details).
These limitations  naturally  lead to the following question: 
\textbf{Can AI models learn transferable knowledge for complex, long-horizon tasks directly from unlabeled real-world videos?} 


\begin{figure}[t]
\includegraphics[width=\linewidth]{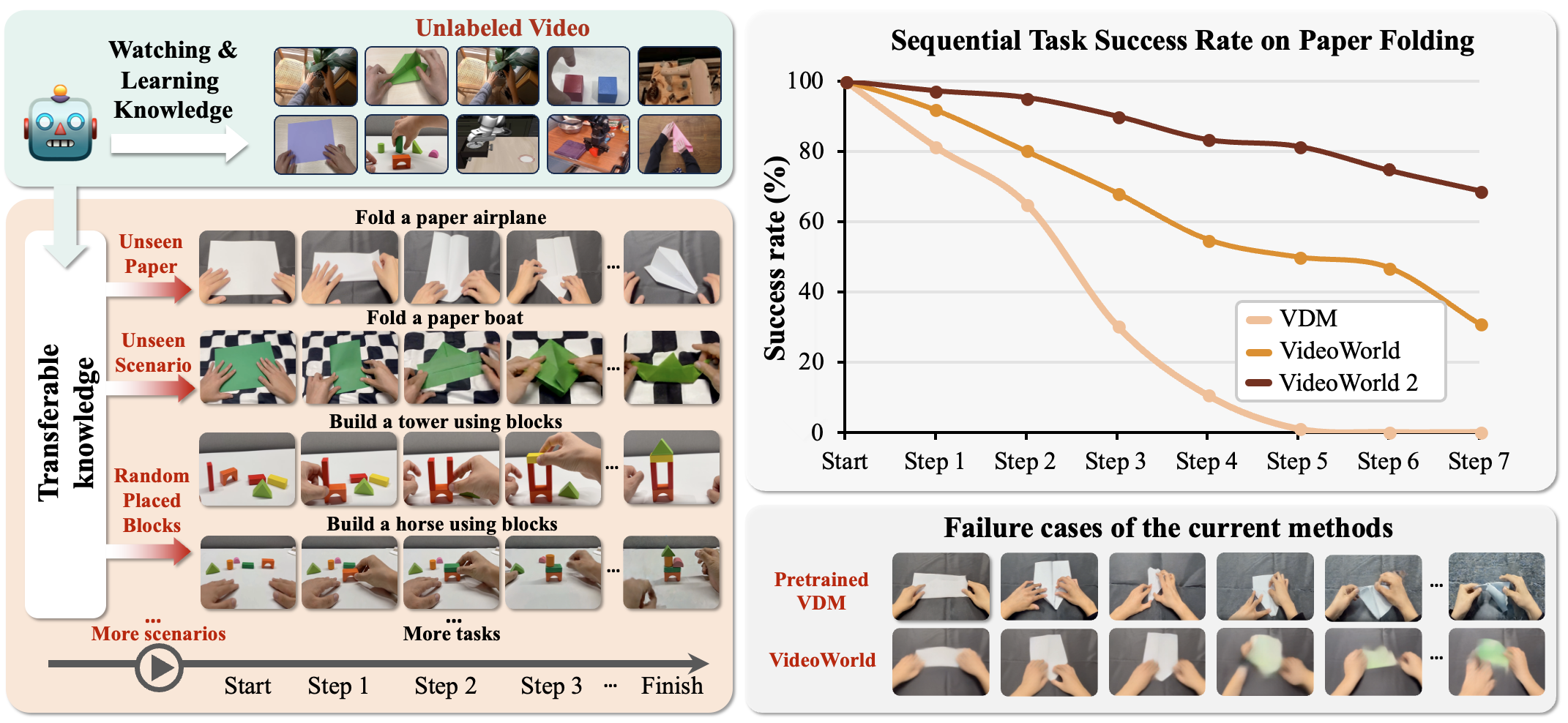}
\centering
\caption{(left) VideoWorld~2 explores how to learn transferable knowledge from unlabeled real-world videos. We construct a handicraft benchmark to evaluate the learned knowledge. (right) Comparison of different frameworks in  success rate for  long-horizon paper folding tasks in Video-CraftBench. We split the task into seven key steps and evaluate sequential success rates (detailed in Sec.~\ref{sec:data}). VDM (e.g., Wan2.2 14B~\cite{wan2025wan}) produces high visual fidelity  but fails to learn task-relevant dynamics or long-horizon policies. 
VideoWorld~\cite{ren2025videoworld} improves policy learning but suffers from poor visual quality in real-world scenarios.  The bottom-right figure presents the failure cases of these baseline methods.
VideoWorld~2 learns more robust latent dynamics while also achieving significantly better visual quality, enabling generalizable long-horizon knowledge learning from videos. 
}
\label{fig:fig1_show}
\vspace{-0.5em}
\end{figure}

In this work, we extend VideoWorld to real-world settings. We first establish a Video-Craft benchmark to systematically evaluate the capabilities of existing   models across two  representative real-world environments. As illustrated in Fig.\ref{fig:fig1_show},  the first environment focuses on \textbf{handicraft making}, which serves as a challenging testbed for learning task knowledge directly from raw videos. These videos involve fine-grained manipulation under diverse environmental appearances, including deformable materials (e.g., paper), viewpoint variation, and frequent occlusions between hands and objects. Moreover, the videos are minute-long and consist of multiple interaction steps, presenting substantially greater complexity and longer temporal horizons than typical entertainment-oriented videos. Second, we study \textbf{robotic manipulation} by learning from  the Open-X dataset~\cite{o2024openx}, which contains real-world demonstration videos, and evaluating on the CALVIN environment~\cite{mees2022calvin}. Since CALVIN features   tasks and visual settings that differ from those in  Open-X, this setup enables a rigorous  assessment of the generalization capability of the learned knowledge. Crucially, learning from these benchmark videos requires   models to extract  task-critical actions from agent motion dynamics and to derive robust policies that can reason and plan correct action sequences across diverse environments.



To address this benchmark, we first   apply VideoWorld directly to real-world videos. Compared to its original training data, these videos exhibit substantially more complex motion dynamics and visual appearances. We observe that VideoWorld struggles to extract task-critical actions and to learn policies generalizable to novel environments. As shown in Fig.~\ref{fig:fig1_show} (bottom left) , the model's predictions contain severe errors, including  distorted hand poses, incorrect object shapes, and inconsistent environmental appearances, ultimately failing to produce accurate and visually coherent action sequences. Moreover, although state-of-the-art video generation models~\cite{wan2025wan,kong2024hunyuanvideo,agarwal2025cosmos} can generate high-fidelity visual details, they similarly fail to faithfully represent   task execution.
We conjecture that these models are unable to  disentangle    task-core actions embedded in visual changes and motion dynamics, and instead overfit to irrelevant visual details. This overfitting  leads to degraded task performance and unstable long-horizon generation. In contrast, humans naturally prioritize essential actions while filtering out extraneous variations.

\begin{figure*}[t]
\includegraphics[width=\linewidth]{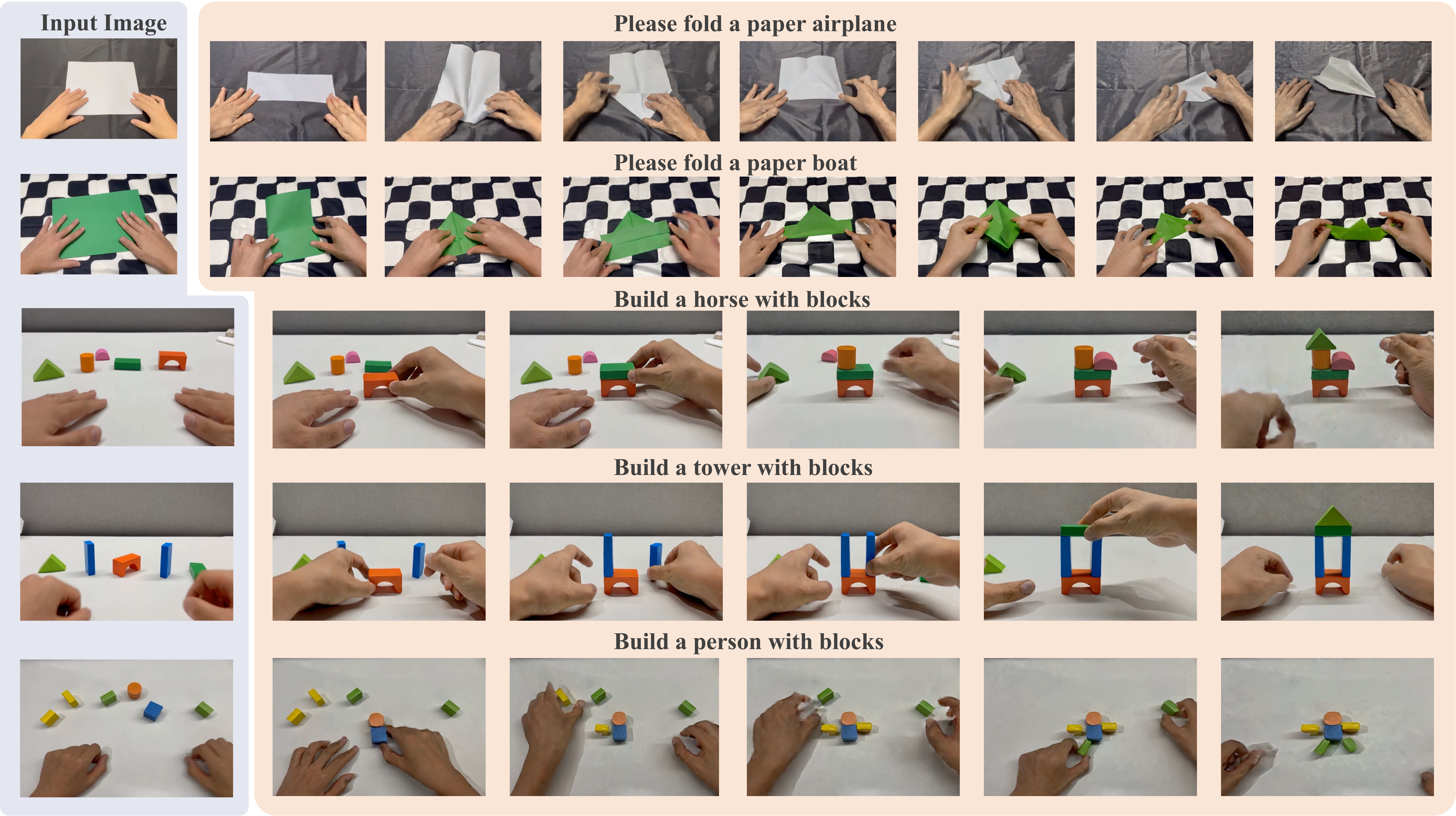}
\centering
\caption{\textbf{Qualitative Results.} VideoWorld~2 learns transferable knowledge and generates long-horizon videos in unseen environments. This figure shows the output on long-horizon handicraft tasks. 
}
\label{fig:res_show}
\vspace{-1em}
\end{figure*}

Inspired by these observations, we propose \textbf{VideoWorld~2}, which aims to robustly acquire knowledge by explicitly  decoupling appearance modeling from action  learning. The key innovation over VideoWorld is the incorporation of prior appearance knowledge during action extraction, which prevents   irrelevant appearance information from being entangled with  the learned action representations. We implement this idea through a novel   \textbf{dynamics-enhanced Latent Dynamics Model (dLDM)}. 
The dLDM comprises a causal VQ-VAE and a pretrained Video Diffusion Model (VDM). The VQ-VAE compresses future visual changes into discrete latent codes that capture task-relevant actions, while the VDM is responsible for modeling  visual appearance, generating high-fidelity reconstructions of task sequences conditioned on these codes. By offloading appearance modeling to the VDM, VideoWorld~2 encourages the latent action codes to focus on concise, semantically meaningful, and transferable actions rather than superficial appearance details. As a result, the learned action policies  are significantly  more robust and generalizable than those produced by VideoWorld. 
As illustrated by the qualitative results in Fig.~\ref{fig:res_show}, VideoWorld~2 successfully executes long-horizon handcraft making tasks in unseen environments, producing coherent task execution videos.

Beyond these qualitative demonstrations, we further compare VideoWorld~2 with prior methods on Video-CraftBench. As shown in Fig.~\ref{fig:fig1_show} (right), VideoWorld~2 excels at long-horizon tasks, outperforming competing methods by learning more generalizable action representations and achieving significantly higher visual quality.
In robotic, pre-training on Open-X further enables VideoWorld~2 to acquire effective manipulation knowledge. This leads to substantial improvements in action prediction performance on the CALVIN environment, demonstrating that the learned actions and policies transfer robustly across diverse robotic tasks.


Our contributions are summarized as follows:
\begin{itemize}
    \item We are the first to investigate learning transferable   knowledge for complex, long-horizon tasks directly from raw real-world videos. Our findings show that disentangling action dynamics from visual appearance modeling is critical for effective knowledge acquisition.
    \item We propose the dynamics-enhanced Latent Dynamics Model (dLDM), which decouples task-relevant dynamics from visual appearance, substantially improving the   quality and transferability of   learned knowledge for both handicraft and robotic manipulation tasks.
    \item We introduce Video-CraftBench, a new benchmark designed to address the largely  unexplored challenge of fine-grained, long-horizon visual reasoning in real-world handicraft tasks. This benchmark provides a foundation for future research on learning transferable knowledge directly from raw videos.
\end{itemize}

\section{Related Work}
\label{sec:related}

\subsection{Video Generation}
Video generation research is dominated by two main paradigms.
First, diffusion-based models set the standard for photorealism, with models like Sora~\cite{openai_sora2}, Veo~\cite{deepmind_veo}, HunyuanVideo~\cite{kong2024hunyuanvideo}, Wan~\cite{wan2025wan}, and CogVideoX~\cite{yang2024cogvideox} generating high-fidelity videos. Second, autoregressive (AR) models, inspired by LLMs~\cite{zhao2023survey_llm}, excel at causal sequence modeling. Works like Lumos-1~\cite{yuan2025lumos}, VideoPoet~\cite{kondratyuk2023videopoet}, and NOVA~\cite{nova_deng2024autoregressive} frame video generation as a next-token prediction task similar to text. Some leading platforms like Cosmos~\cite{agarwal2025cosmos} explore both diffusion and AR branches for comprehensive generation, supporting downstream tasks such as planning~\cite{unipi_du2023learning} and real-time interactivity~\cite{yang2024cogvideox,yu2025survey_ivg}. 
VideoWorld~2 combines the strengths of both paradigms. It leverages the appearance priors of diffusion models to \textit{disentangle} core actions from visual changes, thereby enabling autoregressive modeling of long-horizon policies based on these dynamics.

Furthermore, while some video generation models also explore  ``disentanglement'',  VideoWorld~2 differs substantially in both objective and methodology. {First}, in prior works~\cite{kumar2024trajectory,li2025tokenmotion,bello2025videospats,wang2020g3an,shen2025controllable,xie2020motion,ma2025follow}, disentanglement typically refers to separating motion from appearance for applications like style transfer or visual editing (e.g., camera movement or object-specific changes). In contrast, VideoWorld~2 targets a more demanding objective: reducing task-irrelevant information to learn transferable visual dynamics for complex long-horizon tasks. {Second}, technically, many works~\cite{kumar2024trajectory,li2025tokenmotion,bello2025videospats} rely on explicit geometric supervision to isolate motion signals. Others~\cite{wang2020g3an,shen2025controllable} capture only coarse, global motion semantics rather than temporal dynamics, while methods such as~\cite{xie2020motion,ma2025follow} depend on handcrafted residual encoding or manually separated parameter groups. In contrast, VideoWorld~2 employs a dynamics-enhanced latent dynamics model (dLDM) that suppresses appearance variation and captures task-relevant visual dynamics directly from unlabeled videos, yielding representations suitable for complex task execution, which is beyond the capability of prior approaches.

\subsection{World Models}
World models, which aim to learn physical dynamics for controllable simulation~\cite{survey_wm_defition_ding2025understanding,survey_wm2_zhu2024sora}, are approached from several perspectives. The video generation community~\cite{parker2024genie2,cui2025emu3,pewm,aether,yu2025gamefactory,dreamgen,dreamerv4_hafner2025training} treats them as controllable synthesizers, emphasizing fidelity and consistency. The reinforcement learning/robotics communities~\cite{dreamerv1_hafner2019dream,dreamerv2_hafner2020mastering,dreamerv3_hafner2023mastering,dyn-o_wang2025dyn,harmonydream_ma2023harmonydream,ssvwm_po2025long,dreamweaver_baek2025dreamweaver,EMERALD_burchi2025accurate,S5WM_krinner2025accelerating} view them as learned dynamics simulators for sample-efficient planning. JEPA-style approaches~\cite{vjepa2_assran2025v,vjepa_bardes2024revisiting,flare,dino-wm,dino-world_baldassarre2025back,karypidis2024dino_DINO-Foresight,iwm_garrido2024learning,law_li2024enhancing} avoid pixel-level reconstruction, instead forecasting in an abstract space to benefit downstream tasks.
While these perspectives primarily focus on learning short-term world dynamics for synthesis or planning, VideoWorld~2 addresses the distinct challenge of learning transferable knowledge for complex, long-horizon tasks directly from unlabeled real-world videos.

\subsection{Learning from unlabeled videos}
A key challenge in learning from unlabeled video is extracting meaningful and transferable representations of visual dynamics. While some methods~\cite{xu2023xskill} utilize content-similar video pairs to extract general representations, such paired data is scarce. Some recent works explores fully unsupervised learning of implicit latent actions via strategies like forward-inverse cycle consistency~\cite{lapo_schmidt2023learning,ficc_ye2022become}, VQ-VAE quantization~\cite{lapa_ye2024latent,chen2025moto}, or future frame prediction~\cite{bruce2024genie,adaworld,chen2025villa,tharwat2025latent,chen2024igor}. VideoWorld~2 distinguishes itself in several ways. \textbf{First}, our dLDM models long-horizon visual dynamics, whereas existing latent action models typically focus on short-horizon or pairwise transitions. Moreover, VideoWorld~2 incorporates a pre-trained video generation model to guide the learning of more transferable latent dynamics, while other approaches rely on standard reconstruction decoders. \textbf{Second}, VideoWorld~2 targets minute-long tasks like handicraft involving complex visual dynamics, multiple stages, and substantial appearance variation. This setting is far more challenging than the short-horizon tasks addressed by existing LAMs, such as grasping, toggling, or 2D games~\cite{li2025evaluating,lynch2023interactive,cobbe2020leveraging}.

Regarding VideoWorld~\cite{ren2025videoworld}, it struggles to decouple task-relevant dynamics from visual appearance, leading to appearance drift and motion errors in unseen environments (see Fig.~\ref{fig:supp_fail_vis}). By delegating appearance modeling to the pre-trained VDM, VideoWorld~2 allows its latent space to focus on concise and transferable dynamics rather than appearance details, substantially improving robustness and transferability


\section{Approach}
\label{sec:method}

\begin{figure}[t]
\includegraphics[width=\linewidth]{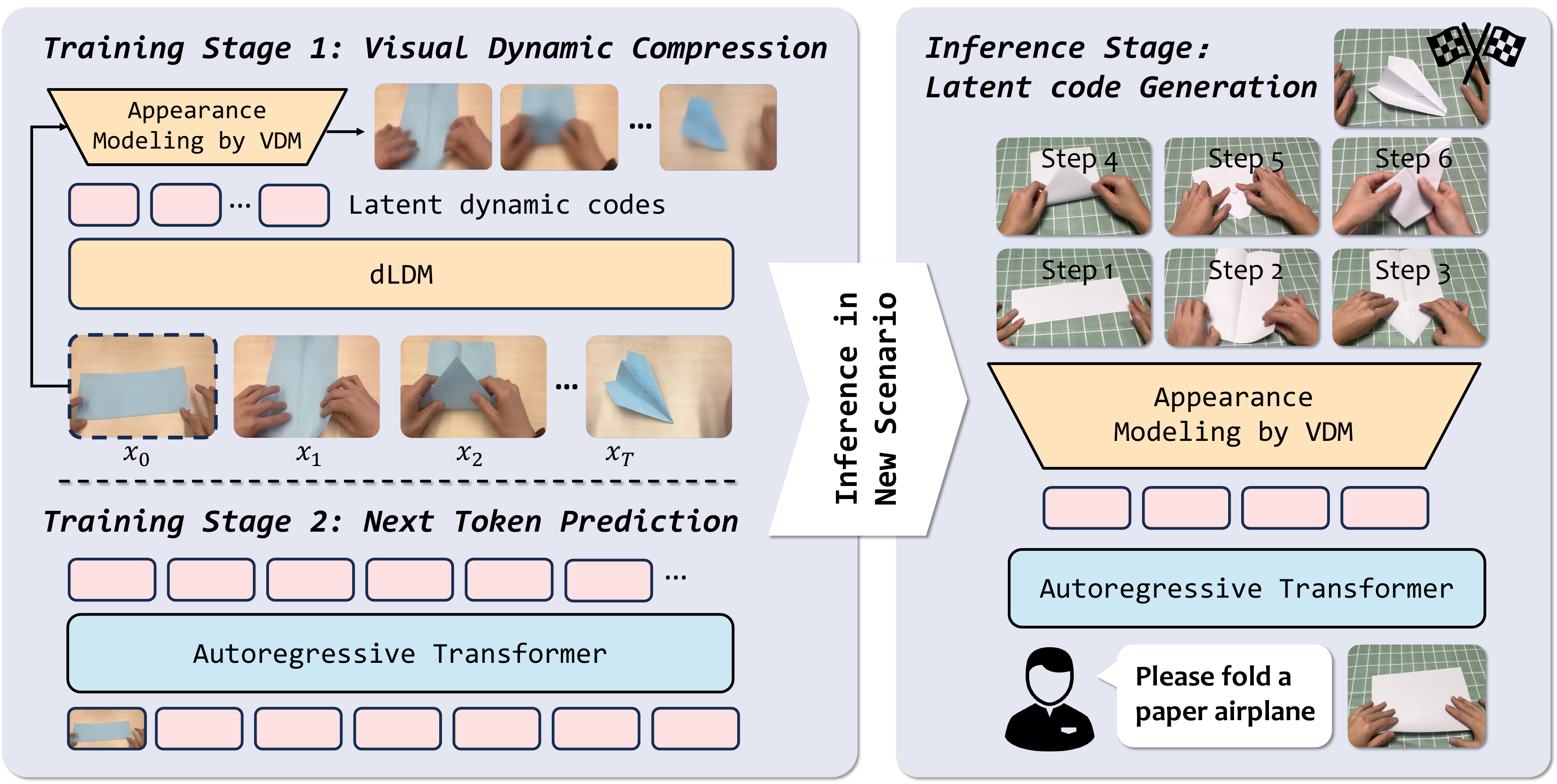}
\centering
\caption{\textbf{Overview of the VideoWorld~2 model architecture.} 
(Left) First, the dLDM compresses future visual changes into compact and generalizable latent codes.  These codes are then modeled by an autoregressive transformer. 
(Right) In inference, the transformer predicts latent codes for a new, unseen environment from the input image, which are subsequently decoded into task execution videos. 
}
\label{fig:method_overview}
\vspace{-1em}
\end{figure}

\subsection{Learning Knowledge from Unlabeled Videos}

\noindent{\textbf{Generative knowledge learning.}} A video can be viewed as a demonstration trajectory that captures world state transitions and the underlying action policy, which constitute the knowledge to be learned. Accordingly, we propose to use a generative model to capture these latent  policies and action dynamics directly from video, without reliance on language supervision. Formally, we define this setting as a tuple  $\mathcal{G}=\left \langle \mathcal{X}, \mathcal{A}, \rho \right \rangle$,  where $\mathcal{X}$ denotes the observation space, $\mathcal{A}$  is the action space, and $\rho$ is  a video generator. Given a sequence of video frames $x\in\mathcal{X}$, our goal is to train $\rho$ to model the conditional distribution of the next frame $x_{t+1}$ given the observation history $x_{0:t}$. This formulation  enables the acquisition of task knowledge through video generation, eliminating the need for explicit labels. 

To transfer the learned knowledge into executable actions for downstream tasks, $\rho$ maps the visual state transitions into action space and thus  serve as a policy model  $\pi(\cdot|x_{0:t}):\mathcal{X}\rightarrow\mathcal{A}$, which predicts actions based on historical observations and executes the tasks accordingly.


\noindent{\textbf{Basic generative framework.}} We adopt mainstream video generation models~\cite{wan2025wan,kong2024hunyuanvideo,agarwal2025cosmos} as the basic framework. They typically utilize a VQ-VAE to encode videos into a compressed representation. The generative process operates within this space to predict future state, which are subsequently decoded back to RGB domain. However, these representations require thousands of discrete tokens or continuous embeddings to capture the full spectrum of visual information, inevitably leading to spatiotemporal redundancy and a sparse distribution of knowledge. As demonstrated in VideoWorld~\cite{ren2025videoworld}, such representations result in an inefficient encoding of the visual changes and motion dynamics tied to critical decisions and actions, thereby hindering the framework from acquiring essential task knowledge from raw videos.



\noindent{\textbf{Latent dynamic model.}}  
To mitigate the above limitation  and facilitate the learning of knowledge, VideoWorld introduces a Latent Dynamic Model (LDM) that compresses future visual changes into a set of compact latent codes, effectively capturing the motion dynamics of multi-step actions.

\begin{figure}[t]
\includegraphics[width=\linewidth]{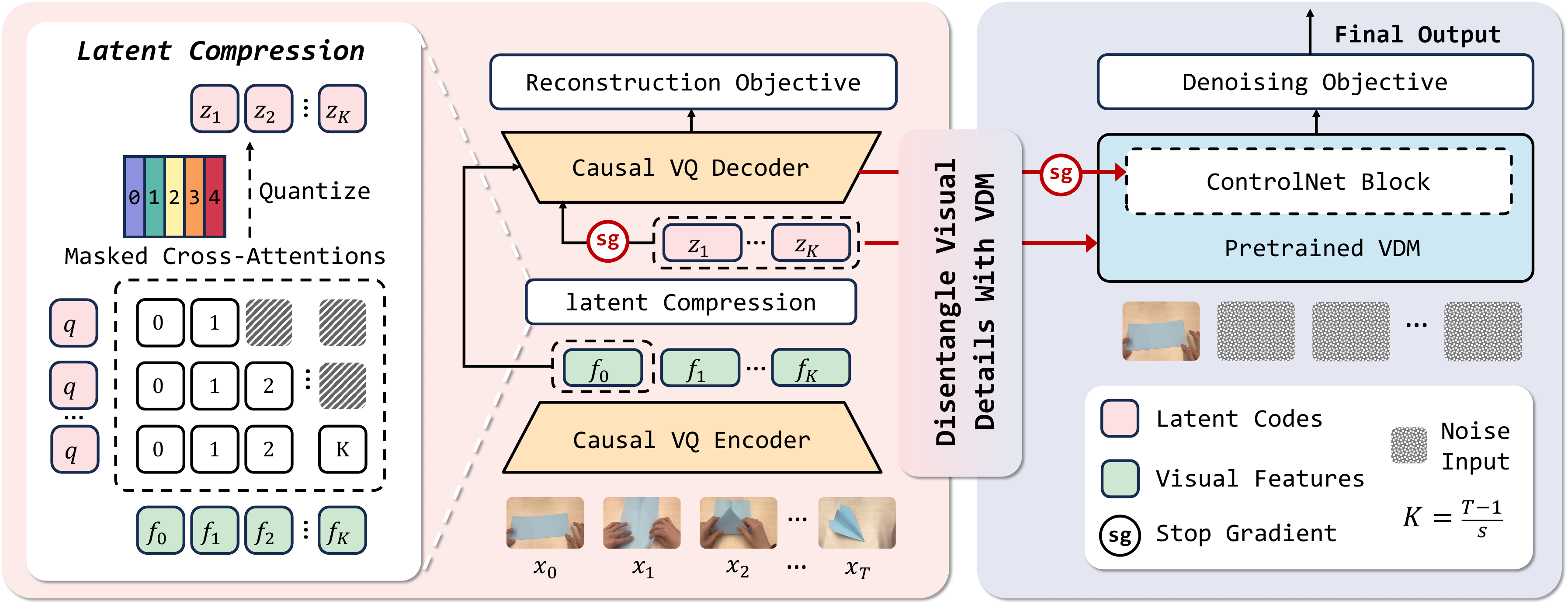}
\centering
\caption{\textbf{The proposed dynamics-enhanced latent dynamics model (dLDM).}  (Left) Latent dynamic model in VideoWorld~\cite{ren2025videoworld}. Visual changes between the first and subsequent frames are compressed into a set of latent codes. (right)  The dLDM proposed in VideoWorld 2. It employs a pre-trained VDM as an appearance prior, yielding better latent codes and facilitating high-fidelity video output.
}
\label{fig:ldm_comp}
\vspace{-1em}
\end{figure}

Specifically, as shown in the left panel of Fig~\ref{fig:ldm_comp}, the LDM utilizes a MAGVITv2-style~\cite{yu2024language} causal codec. The encoder first maps an input clip $x$ of length $T$ to a feature sequence $f_{0:K}$, where $K=1+\lfloor\frac{{T-1}}{s}\rfloor$ and $s$ is the temporal downsampling stride. Next, it defines $N$ learnable query embeddings $q=\{q^n\}_{n=1}^N$. As shown in Fig.~\ref{fig:ldm_comp} (left), these queries use cross-attention to capture change information in $\{f_{0:k}\}_{k=1}^K$, yielding a continuous representation $z=\{z_k^n\}_{k=1, n=1}^{K, N}$. This representation is then quantized to prevent the LDM from learning shortcuts (e.g., trivially copying $f_k$ to $z_k$). Finally, the decoder uses $f_0$ and the quantized $z$ to reconstruct the subsequent frames in a causal manner. The training objective is to minimize the $\ell_2$ distance between the original and reconstructed frames. We refer to these embeddings as latent dynamics codes. 

\begin{figure}[t]
\includegraphics[width=\linewidth]{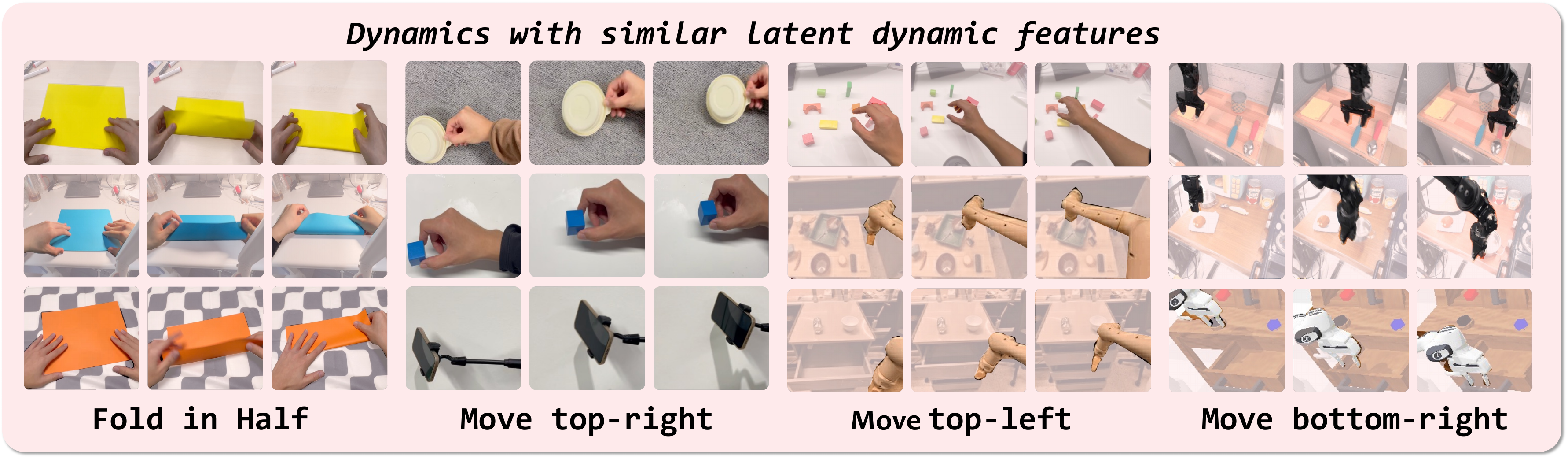}
\centering
\caption{\textbf{Video clips with similar latent dynamic features.} The text below represents the dynamic type.}
\label{fig:method_overview}
\vspace{-1em}
\end{figure}

\subsection{Dynamics-enhanced Latent Dynamic Model} While VideoWorld  effectively handles long-term reasoning in synthetic environments by encapsulating temporal dynamics into compact latent codes, it still struggles with real-world application. As shown in the bottom of Fig.~\ref{fig:supp_fail_vis}, the model fails to generate coherent long-horizon sequences or generalize to new environments. Specifically, applying the latent dynamics learned from handcrafting training videos to novel environments with different desktops, paper materials and arms, frequently results in significant scene drift and erroneous actions.
We also evaluate several state-of-the-art video generation models~\cite{wan2025wan,kong2024hunyuanvideo,agarwal2025cosmos}. As shown in Sec.~\ref{subsec:res_basic} and Fig.~\ref{fig:supp_fail_vis}, although these models can produce high-fidelity frames, they also fail to reproduce fine-grained, long-horizon skills in new contexts, even when detailed language instructions are provided.

\noindent{\textbf{Entanglement of dynamics and appearance.}
We conjecture that these limitations stem from the insufficient disentanglement of action dynamics and visual appearance. In the basic framework, the joint modeling of these components impedes the effective extraction of task-core action knowledge. Regarding VideoWorld, we find its learn latent codes capture irrelevant visual details{\textemdash}such as background motion, lighting changes, textures, and camera displacement{\textemdash}making the model sensitive to environmental changes, as further discussed in Sec.~\ref{subsec:alb}. Consequently, this incomplete separation ultimately leads to poor consistency in long-horizon tasks and limits the model's ability to generalize to new environments.

\noindent{\textbf{Dynamics-enhanced latent dynamic model.}} 
To address this issue, we introduce a dynamics-enhanced Latent Dynamic Model (dLDM). The key mechanism is to replace the original LDM decoder with a pretrained Video Diffusion Model (VDM). By leveraging the capacity of VDM for high-fidelity reconstruction to handle the appearance modeling, we allow the LDM encoder and learnable query embeddings to focus exclusively on capturing task-relevant visual changes. Although the VDM does not contain any knowledge of the target task’s dynamics, it is highly effective at producing realistic visual content once given appropriate dynamics guidance, making it well-suited for this decomposition (As shown in Fig.~\ref{fig:res_show} and Fig.~\ref{fig:umap}). 

Concretely, as shown in the right panel of Fig.~\ref{fig:ldm_comp}, the dLDM consists of a causal VQ-VAE that encodes future visual changes into discrete latent codes and a pretrained VDM that reconstructs high-fidelity frames. The latent codes are provided to the VDM through a projection layer and causal cross-attention. Since the VDM handles appearance modeling, the latent codes are relieved from encoding fine-grained visual details and can instead focus on capturing task-relevant dynamics. To maintain temporal correctness, we enforce causal attention in the VDM so that features at time $t$ attend only to information up to time $t$. 

Directly training the VDM to generate future frames from noise would be extremely slow and prone to incorrect motion, as it has never been trained on the target tasks like long-horizon handicraft making. Therefore, we reuse the VQ-VAE decoder to reconstruct latent codes into low-fidelity, motion-rich outputs, providing coarse temporal cues such as hand movements and object displacements. This signal is fed into the VDM via a gradient-stopped, ControlNet-like~\cite{zhang2023control} branch. This stabilizes training and allows the VDM to focus on refining appearance rather than inferring motion from scratch. Additionally, we stop the gradient flow of the decoder to the latent codes to prevent the introduction of irrelevant noise. We ablate this design in Sec.~\ref{subsec:alb}.

By leveraging the VDM, the learned latent codes are much less affected by changes in appearance and stay consistent across environments. As shown in Sec.~\ref{fig:umap}, latent codes corresponding to the similar action exhibit tighter intra-class alignment and reduced cross-environment variance, indicating robust and transferable dynamics.

\noindent{\textbf{Auto-regressive transformer.}}
After extracting the latent codes, we use an autoregressive transformer to model the latent dynamics sequence. For each video $x_{0:T}$, the dLDM extracts latent codes $\{z_k^n\}_{k=1,n=1}^{K,N}$. We flatten these into a sequence and train the transformer to predict them, conditioned on the initial frame $x_0$ and the task instruction. This allows the model to learn the long-term patterns in complex tasks.

During inference (Fig.~\ref{fig:method_overview}, right), given a single input frame from a new, unseen environment, the transformer predicts future latent dynamics based on its learned task representation, and the dLDM decodes them into coherent long-horizon execution videos. This allows VideoWorld~2 to transfer its learned dynamics to new environments and execute extended action sequences beyond those observed during training.

\section{Video-CraftBench}
\label{sec:data}

\begin{figure*}[t]
\includegraphics[width=\linewidth]{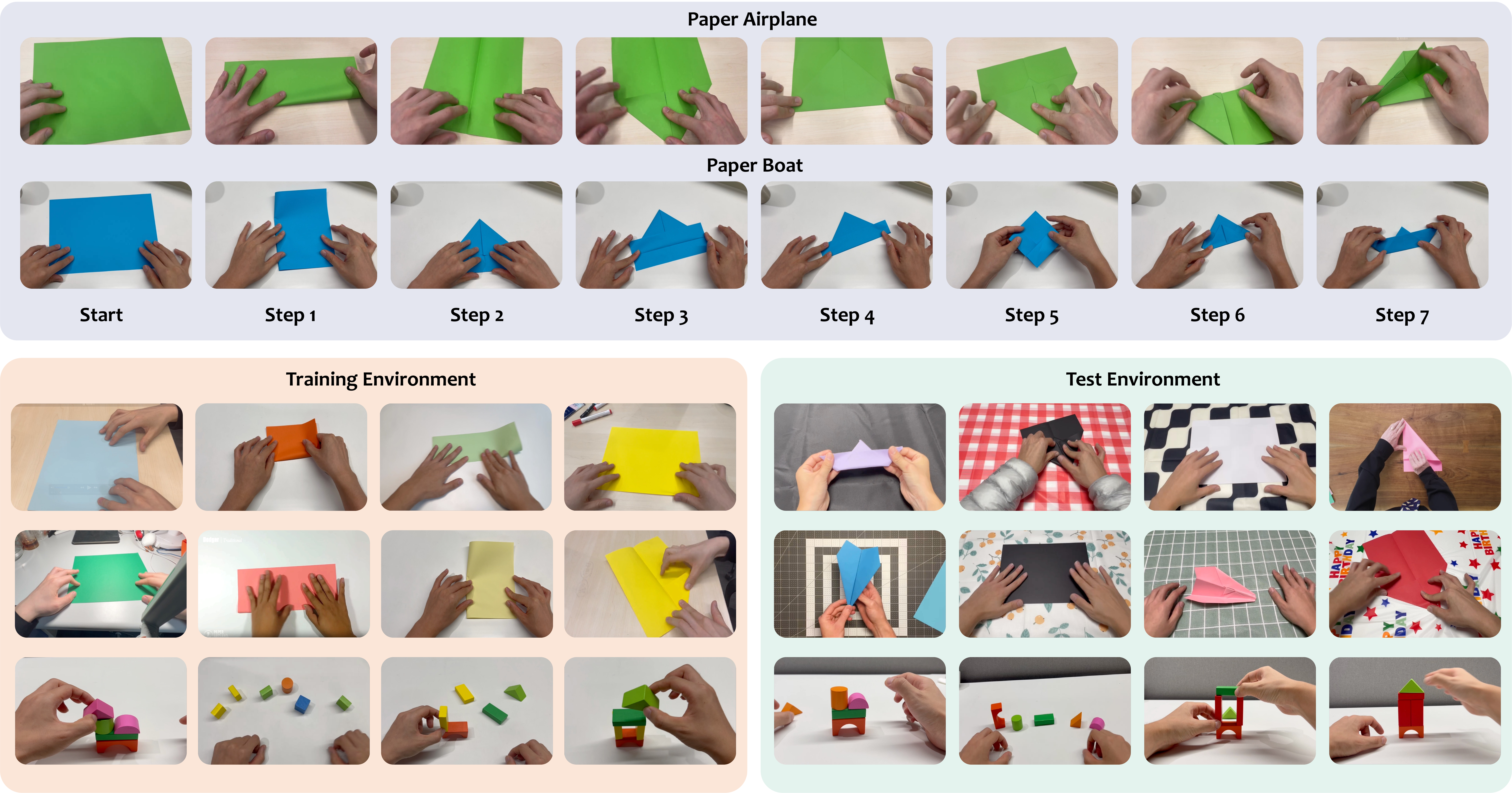}
\centering
\vspace{-0.5em}
\caption{\textbf{Overview of Video-CraftBench} and keys steps of paper folding tasks. Best viewed in color.
}
\label{fig:supp_key_step}
\vspace{-1em}
\end{figure*}

Our objective is to learn transferable knowledge for complex and long-horizon tasks from real-world videos. We prioritize tasks that demand multi-step planning and feature delicate manipulations. To this end, we introduce the Video-CraftBench, a dataset of first-person video tutorials covering five long-horizon handcraft tasks: folding a paper airplane, folding a paper boat, and building a tower/horse/person using blocks.

\subsection{Dataset Generation}
These handcraft tasks feature fine-grained, visually diverse manipulations that are difficult to articulate with language. We collected $\sim$7 hours (split into $\sim$9.5k clips) of tutorial videos via manual recording and internet sources, enabling long-horizon knowledge evaluation. The paper folding tasks typically last 40-80 seconds and block-building tasks last 20-30 seconds. The benchmark is split into training and test sets. The test set ($\sim$150 separately collected videos) features novel backgrounds, paper textures, and block arrangements not seen in training to assess knowledge transferability. Supp.~\ref{sec:supp_data} provides more details.

\subsection{Evaluation}
\label{subsec:vidcraft_eva}
Our evaluation focuses on two main aspects: the accuracy of key actions and the visual quality of the generated videos. Accordingly, we use the following metrics for evaluation:
\begin{itemize}
    \item  \textbf{Sequential task success rate}: We decompose paper folding into 7 key steps (Fig.~\ref{fig:supp_key_step}) and train a DINOv2-based classifier~\cite{oquab2024dinov2} to detect their completion.  The classifier is trained to evaluate only action correctness, disregarding appearance consistency (e.g., appearance drift from the initial frame). To train it, we sample and annotate frames from the training set, test set, and manually-verified generated videos, labeling each as a specific step or ``failed''. For evaluation, we generate 3 video rollouts per test case conditioned on the first frame. A step is only successful if all preceding steps are complete, allowing us to assess long-horizon knowledge acquisition. For the shorter block-building tasks, a similar classifier is trained, but only to verify the final generated state.

    \item   \textbf{Visual quality}: We also assess visual quality in the test environment, using standard LPIPS~\cite{zhang2018lpips} and SSIM~\cite{ssim} metrics to quantify visual fidelity and content consistency with the input.

\end{itemize}

\section{Experiments}
\subsection{Implementation Details}
We employ the NVIDIA Cosmos AR 4B model~\cite{agarwal2025cosmos} as our auto-regressive transformer, repurposing its next-token prediction capability to predict our latent codes. For the appearance prior, we utilize the Cosmos DiT 2B model~\cite{agarwal2025cosmos}, leveraging its high-fidelity I2V capabilities to generate 93-frame ($\sim$5s at 16 fps), 480px videos.
The dLDM also processes a 93-frame clip at a time by default, using a vocabulary size of 1000 (FSQ levels [8, 5, 5, 5]) and an embedding length $N=4$.  To improve training efficiency, dLDM first applies a short warm-up where the latent codes are optimized solely using the original reconstruction objective. This warm-up is similar to the training strategy of the original VideoWorld LDM, enabling the latent codes to rapidly learn to compress visual changes and motion dynamics, while allowing the decoder to reconstruct low-fidelity video clips containing agent motion trajectories based on the initial frame and codes. Consequently, when we switch to the disentangled scheme, the decoder can provide robust motion conditioning to stabilize training. The trainable parameters in our framework include the AR transformer, the dLDM autoencoder, the DiT, and the projection layer mapping the latent codes to the DiT.


\subsection{Benchmarks and Baselines}
\label{subsec:res_basic}

\noindent{\textbf{Benchmarks.}}
We evaluate VideoWorld~2 on two benchmarks: Video-CraftBench assesses learning from long-horizon, real-world tasks with fine-grained actions. 
For robotics, we train our latent representation on the large-scale OpenX dataset and test its knowledge transfer to CALVIN. CALVIN features 34 tasks and uses a long-horizon, sequential evaluation protocol similar to our paper folding setup: models must complete a 5-task sequence, where success is contingent on all preceding tasks.

\noindent{\textbf{Baselines.}} To demonstrate the effectiveness of VideoWorld~2, we establish three strong baselines for comparative analysis on the aforementioned benchmarks:
\begin{itemize}
\item Pre-trained video generation model: We select four state-of-the-art video generation models: NVIDIA Cosmos AR 4B~\cite{agarwal2025cosmos}, NVIDIA Cosmos DiT 2B~\cite{agarwal2025cosmos}, Wan2.2 14B~\cite{wan2025wan}, and HunyuanVideo 13B~\cite{kong2024hunyuanvideo}. We fine-tune them on Video-CraftBench. To better align with their original training, we also provide detailed language annotations for each key step using Qwen2.5-VL 72B~\cite{bai2025qwen2} to aid generation. During inference, these models generate task execution videos conditioned on an initial image and text instructions. They then generate subsequent video clips autoregressively, based on their own previous output.

\item Latent action models: We compare our approach with five concurrent works~\cite{chen2025moto, lapa_ye2024latent, adaworld, wang2025coevolvinglatentactionworld, wu2024ivideogpt} that leverage latent action models. These methods compress inter-frame visual changes into latent codes, serving as a pre-training objective for manipulation tasks while allowing for video reconstruction. Consequently, we apply their latent extraction methodologies to our benchmark to assess their ability to extract knowledge from long-horizon real-world videos. 

\item VideoWorld~\cite{ren2025videoworld}: This is the most relevant baseline to our work. It uses the original latent dynamic model for video knowledge learning.

\end{itemize}

\subsection{Results on Video-CraftBench}
\label{subsec:res_ldm}

\noindent{\textbf{Training with Video-CraftBench only.}} 
In Tab.~\ref{table:paper}, we first evaluate pre-trained video generation models fine-tuned on Video-CraftBench (row 1-4). While these models achieve high success rates ($>$68\%) in the first step of the paper folding task and up to 38\% in the block stacking task, their performance deteriorates rapidly for complete, minute-long sequences.
By step 4 of paper folding (Column 7), success rates drop to $\leq$10.6\%, with a total failure to generate subsequent steps. Crucially, this failure persists even though we have provided detailed textual descriptions for each step during training and inference to leverage their text-conditioning capabilities. This demonstrates that current models struggle to learn complex, long-horizon knowledge from real-world videos, underscoring the need for a more effective paradigm.

Subsequently, we evaluate existing methods that leverage latent action or dynamic models.  Following their protocols, we train these baselines on Video-CraftBench and train our AR transformer to predict their latent codes.  Note that due to LAPA~\cite{lapa_ye2024latent}'s structural constraints, decoding its codes into long-horizon sequences causes severe degradation. Thus, we mark its results as ``N.A.''. For the other three baselines, Moto~\cite{chen2025moto} employs a pre-trained vision encoder to extract dynamics, AdaWorld~\cite{adaworld} utilizes an auxiliary diffusion head for video synthesis, and VideoWorld~\cite{ren2025videoworld} features capabilities for long-horizon dynamics compression and generation. Although they generate coherent videos, they fail to generalize to novel environments. Specifically, none successfully completed the full paper folding sequence under new settings (e.g., different desktops or paper styles). As shown in Fig.~\ref{fig:supp_fail_vis}, their predictions exhibit substantial action errors and scene distortions, indicating that their latent codes overfit to irrelevant appearance information.

In contrast, VideoWorld 2 (row 9) generates complete and coherent task sequences in test environments. Remarkably, without requiring large-scale pre-training for these latent codes, training solely on Video-CraftBench achieves a success rate of 68.8\% on the paper folding task and up to 81.5\% on the block stacking task. This demonstrates that our dLDM efficiently extracts core task-relevant action information while filtering out extraneous details like background clutter, object variations, and camera noise. Consequently, it generalizes effectively to unseen environments. Furthermore, benefiting from the appearance priors of the VDM, VideoWorld~2 not only generates accurate actions but also produces videos with higher SSIM and PSNR metrics. Fig.~\ref{fig:res_show} and Supp.~\ref{sec:supp_vis} visualize our generated sequences.

\begin{table}[t]

  \scriptsize 
  \centering
  \renewcommand{\arraystretch}{1.1} 
  \setlength{\tabcolsep}{5pt} 
  \begin{tabular}{p{2.3cm}c ccccccc ccc cc}
   \toprule
   \multirow{2}*{Method} &\multirow{2}*{Fine-tuning} &\multicolumn{7}{c}{Sequential Paper folding Success Rate (\%) } &\multicolumn{3}{c}{Block Success Rate (\%)} &\multicolumn{2}{c}{Visual Quality}
   \\
   \cmidrule(lr){3-9} \cmidrule(lr){10-12} \cmidrule(lr){13-14}
  & &1 &2 &3 &4 &5 &6 &7 &Human &Tower &Horse &SSIM$\uparrow$ &LPIPS$\downarrow$\\
   \midrule
   \multicolumn{13}{l}{{\color{gray}\textit{Pre-trained Video Generation Model}}} \\
   Cosmos AR 4B~\cite{agarwal2025cosmos} & Craft-text & 68.4 & 56.7 & 11.5 & 3.3 & 0.0 & 0.0 & 0.0  &10.1 &18.0 &12.0 &0.643 &0.312 \\
  Cosmos DiT 2B~\cite{agarwal2025cosmos} & Craft-text & 73.4 & 63.3 & 20.0 & 6.7 & 0.0 & 0.0 & 0.0  &24.2 &21.3 &19.7 &0.680 &0.264 \\
   Hunyuan-13B~\cite{kong2024hunyuanvideo} & Craft-text & 76.9 & 68.1 & 27.5 & 5.8 & 0.0 & 0.0 & 0.0 &30.9 &38.4 &31.5 &0.703 &0.255\\
  Wan 2.2 14B~\cite{wan2025wan} & Craft-text & 81.2 & 75.0 & 30.4 & 10.6 & 0.0 & 0.0 & 0.0 &39.7 &42.6 &34.1 &0.719 &0.237 \\
   \midrule
   \multicolumn{13}{l}{{\color{gray}\textit{Learning with Latent Action/Dynamic Model}}} \\
   LAPA~\cite{lapa_ye2024latent}  &Craft  &\multicolumn{7}{c}{N.A.} &\multicolumn{3}{c}{N.A.} &\multicolumn{2}{c}{N.A.} \\
  Moto~\cite{chen2025moto}  &Craft 
   &19.1 &11.7 &3.3 &0.0 &0.0 &0.0 &0.0 &11.5 &10.1 &9.8 &0.585 &0.394
   \\
AdaWorld~\cite{lapa_ye2024latent}  &Craft  
   &43.6 &39.8 &27.4 &10.8 &0.0 &0.0 &0.0 &20.7 &13.1 &15.0 &0.611 &0.378
   \\

VideoWorld~\cite{ren2025videoworld}  & Craft  &\underline{70.3}  &\underline{66.7} &\underline{42.5}  & \underline{21.3} & \underline{6.7} & 0.0 & 0.0  &\underline{23.8} &\underline{33.9} &\underline{27.8} &\underline{0.680} &\underline{0.351}\\
VideoWorld 2 & Craft   &{\bfseries 97.2}  &{\bfseries 95.3} &{\bfseries 90.0} &{\bfseries 83.3} &{\bfseries 81.4} &{\bfseries 74.6} &{\bfseries 68.8}   &{\bfseries 70.0} &{\bfseries 81.5} &{\bfseries 80.9} &{\bfseries 0.770} &{\bfseries 0.205}\\
   \midrule
   iVideoGPT~\cite{wu2024ivideogpt} &OpenX \& Craft &23.1 &18.7 &13.3  &3.7 &0.0 &0.0 &0.0  &15.3 &11.0 &12.6 &0.588 &0.390 \\
  Moto~\cite{chen2025moto}  &OpenX \& Craft 
   &43.1 &35.3 &30.7 &25.5 &18.3 &9.7 &0.0 &17.4 &15.3 &16.0 &0.596 &0.387
   \\
   AdaWorld~\cite{lapa_ye2024latent}  &OpenX \& Craft   &49.5 &41.6 &34.8 &30.7 &22.3 &19.8 &13.0 &37.4 &29.8 &29.1 &0.624 &0.365 \\

   CoLA~\cite{wang2025coevolvinglatentactionworld}  &OpenX \& Craft   &83.5 &74.4 &\underline{69.1} &\underline{64.8} &\underline{52.3} &\underline{49.8} &\underline{40.2} &\underline{54.1} &52.4 &\underline{49.9} &\underline{0.668} &\underline{0.289} \\
   
   VideoWorld~\cite{ren2025videoworld}  &OpenX \& Craft &\underline{91.7}  &\underline{75.0} &68.2 &63.1 &51.7 &48.2 &31.9   &47.3 &\underline{52.7} &49.8 &0.601 &0.389\\
   VideoWorld 2 & OpenX \& Craft &{\bfseries 98.2}  &{\bfseries 96.4} &{\bfseries 90.1} &{\bfseries 86.7} &{\bfseries 83.3} &{\bfseries 81.7} &{\bfseries 72.3} &{\bfseries 74.0} &{\bfseries 83.0} &{\bfseries 85.8} &{\bfseries 0.774} &{\bfseries 0.193}\\

   \bottomrule
  \end{tabular} 
\vspace{-0.5em}
\caption{\textbf{Comparison on Video-CraftBench.} Craft-text denotes using step-by-step textual description.}
\label{table:paper}
\vspace{-1em}
\end{table}

\noindent{\textbf{Data scaling with OpenX.}} Next, we evaluate whether the latent codes of VideoWorld 2 can benefit from larger-scale data. To this end, we incorporate the OpenX dataset~\cite{o2024openx}, which contains a vast collection of manipulation demonstrations. Compared to Video-CraftBench, OpenX features a diverse array of robotic agents and environmental appearances, making it an ideal testbed for assessing VideoWorld 2's ability to filter out agent-specific and environmental factors while scaling its action extraction capabilities. 

Specifically, we first train dLDM and baseline models on a combination of OpenX and Video-CraftBench, followed by training the AR transformer exclusively on Video-CraftBench using the resulting latent codes. As shown in line 10-12 of Tab.~\ref{table:paper},  the baselines show improved quality, raising success rates across tasks. However, Moto and iVideoGPT still fail to complete the full folding sequence. While VideoWorld can generate complete sequences, its final-step success rate remains low at 31.9\%. We also examine CoLA~\cite{wang2025coevolvinglatentactionworld}, a concurrent work that also uses a VDM to optimize latent action codes. However, it is limited to short 2-frame transitions and ignores the structured temporal cues from coarse VAE outputs. Adopting CoLA's training scheme yields a limited final-step success rate of 40.2\%. 
In contrast, VideoWorld~2 further improves upon its strong baseline performance, achieving a final-step success rate of 72.3\% on paper folding and up to 85.8\% on block stacking. As shown in the right panel of Fig.~\ref{fig:method_overview}, similar latent codes in OpenX and Video-CraftBench correspond to videos with similar motion patterns despite differing environments and agents, demonstrating our transferability across diverse agents and settings.

\subsection{Results on CALVIN}
\label{subsec:res_calvin}
\noindent{\textbf{In-domain latent pre-training on CALVIN.}} To demonstrate that our latent codes aid in learning manipulation skills, we conduct a latent pre-training experiment, following LAPA. We first pre-train the AR transformer on latent codes from 22k CALVIN trajectories, then fine-tune it on only 2k ground-truth action labels. As shown in Tab.~\ref{table:calvin_wide}, this pre-training strategy significantly boosts VideoWorld~2's performance (Idx 4) on long-horizon tasks compared to the baseline (Idx 2). Notably, its performance approaches the model trained on the 22k full action label dataset (Idx 1), demonstrating high data efficiency. While LAPA (Idx 3) also benefits from this protocol, it still lags behind VideoWorld~2, highlighting the effectiveness of our latent codes.

\noindent{\textbf{Cross-domain latent pre-training on CALVIN.}} We further evaluate transferability by pre-training on the 1.3M OpenX dataset and fine-tuning on 22k CALVIN trajectories. As shown in Tab.~\ref{table:calvin_wide}, this pre-training significantly boosts success rates for LAPA (Idx 6) and VideoWorld~2 (Idx 7) compared to training only on labeled CALVIN data (Idx 1). Consistent with previous findings, VideoWorld~2 benefits more from this pre-training on long-horizon tasks. We also include a video next-token prediction baseline pre-trained on OpenX (Idx 5). The results show that our latent pre-training is more effective, validating it as a more efficient knowledge transfer paradigm than pre-training directly on videos. Supp.~\ref{sec:supp_detail} provides more training details.

\subsection{Ablation Study}
\label{subsec:alb}





 

\begin{table*}[t]
   \footnotesize 
  \centering 
  \setlength{\tabcolsep}{0pt}
  \begin{tabular*}{\linewidth}{@{\extracolsep{\fill}} c l c c c c c c c c c}
   \toprule
   
   \multirow{2}*{Idx} &\multirow{2}*{Method} &\multirow{2}*{\makecell[c]{Pretraining\\Type}} &\multirow{2}*{Pretraining}  &\multirow{2}*{Fine-tuning} &\multicolumn{6}{c}{Sequential Task Success Rate (\%) } \\
   \cmidrule(lr){6-11} 
   & & & & &1 &2 &3 &4 &5 & Avg. Len. \\
   \midrule
    \multicolumn{11}{l}{{\color{gray}\textit{In-Domain Latent Pretraining}}} \\ 
    1&Transformer (Oracle) &- & - & ABCD${\rightarrow}$D & 80.9 & 55.6 &44.5 & 31.3 & 24.6 &2.36 \\
    2&Transformer (Oracle) &- & - & 10\% data &50.5 & 35.4 &20.1 & 5.20 &0 &1.11 \\

    3&LAPA~\cite{lapa_ye2024latent} &Latent & ABCD${\rightarrow}$D & 10\% data & 74.4 & 45.8 & 25.2 & 15.3 & 2.30 &1.49 \\
    4&VideoWorld~2 &Latent & ABCD${\rightarrow}$D & 10\% data &75.8 &47.9 &31.8 &20.4 &9.70 &1.87   \\
    \midrule
    \multicolumn{11}{l}{{\color{gray}\textit{Cross-Domain Latent Pretraining}}} \\ 
    5&Transformer (Oracle) &Video & OpenX & ABCD${\rightarrow}$D  &{\bfseries 85.9} &\underline{60.4} &46.0 & 30.7 & 23.0 &2.46\\
    6&LAPA~\cite{lapa_ye2024latent} &Latent & OpenX & ABCD${\rightarrow}$D &84.0 &58.8 &\underline{46.2} &\underline{35.4} &\underline{27.0} &\underline{2.51} \\
    7&VideoWorld~2 &Latent & OpenX & ABCD${\rightarrow}$D &\underline{88.5} &{\bfseries 64.6} &{\bfseries 55.8} &{\bfseries 47.5} &{\bfseries 30.9} &{\bfseries 2.88}   \\
   \bottomrule
  \end{tabular*}
\caption{{\bfseries Comparison on CALVIN benchmark.} ``10\% data'' means 2k trajectories randomly sampled from the ABCD${\rightarrow}$D training set. ``Oracle'' means directly training the transformer using ground-truth action label as supervision.}
\vspace{-1em}
\label{table:calvin_wide}
\end{table*}

\noindent{\textbf{Code transferability is enhanced by VDM.}} 
In Tab.~\ref{tab:abla_arch}, we first ablate the introduction of the VDM prior (line 1 vs. 2). We find that incorporating the VDM enhances codes transferability and visual quality, yielding a $\sim$30\% increase in success rate and an improved LPIPS score. This result validates the effectiveness of our appearance-dynamic decoupling design.  
To provide a more intuitive illustration, we visualize the latent codes using UMAP in Fig.~\ref{fig:umap}. We randomly sample 4000 trajectories from CALVIN and Bridge (included in OpenX)~\cite{walke2023bridgedata} and label each by its robot arm action (up, down, left, or right). As shown in Fig.~\ref{fig:umap} (left), with VDM, latent codes for the same action align much more tightly across environments, indicating more consistent and transferable dynamics. In contrast, without VDM (Fig.~\ref{fig:umap}, right), the codes separate strongly by environment, revealing a loss of cross-environment consistency.




\noindent{\textbf{Effect of the original decoder.}} 
In Sec.\ref{sec:method}, we first warm up the dLDM with the original decoder during training. In subsequent stages, we discard the reconstruction loss from the VQ-VAE decoder to prevent noise injection. However, we observe that after warm-up, the decoder can reconstruct latent codes into videos that preserve coherent object motion (e.g., hand movements and displacements), although visual details remain blurry. Consequently, we utilize these reconstructions as conditions for the VDM via a ControlNet-like structure to enhance output quality and stabilize training. Tab.\ref{tab:abla_arch} further investigates this decoder.

First, we stop the gradient from the decoder without using the reconstructed video (row 3) to verify if this decoder introduces noise. Compared to row 2, this yields a $\sim$20\% improvement in success rate, suggesting that the original decoder indeed introduces extraneous noise that degrades latent representation performance. Second, we investigate the utility of the reconstructed video itself. We find that utilizing this motion conditioning (rows 4 and 5) stabilizes video output, yielding a substantial improvement of $\sim$0.9 in LPIPS and up to $\sim$20\% in task success rate. This benefit is more pronounced in the longer paper folding task compared to block stacking, demonstrating the effectiveness of our method for long-horizon video generation.



\begin{table*}[t]
\centering
\hfill
\subfloat[
dLDM architecture.
\label{tab:abla_arch}
]{
\centering
\begin{minipage}[t]{0.37\linewidth} 
\centering
\tablestyle{1.5pt}{1.05} 
\scriptsize
\begin{tabular}{c c c c cc}
    \multicolumn{1}{c}{\multirow{2}*{\makecell{Pre-trained \\ VDM }}}
      &\multicolumn{1}{c}{\multirow{2}*{\makecell{Decoder \\ Stop-Grad}}} &\multicolumn{1}{c|}{\multirow{2}*{\makecell{Ctrl- \\ Net}}} &\multicolumn{3}{c}{Video-CraftBench}    \\
     & & \multicolumn{1}{c|}{} &Paper &Block & LPIPS$\downarrow$   \\
    \shline
    \specialrule{0em}{0pt}{1pt}
    & &\multicolumn{1}{c|}{} &0.0 &28.5 &0.312\\
    \small{\checkmark} & &\multicolumn{1}{c|}{} &30.3 &45.2 &0.297 \\
    \small{\checkmark} &\small{\checkmark} & \multicolumn{1}{c|}{} &47.3 &\underline{54.7} &0.275 \\
    \small{\checkmark} & & \multicolumn{1}{c|}{\small{\checkmark}} &\underline{51.1} &52.0 &\underline{0.213} \\
    \small{\checkmark} & \small{\checkmark} & \multicolumn{1}{c|}{\small{\checkmark}} &{\bfseries 68.8} &{\bfseries 77.5} &{\bfseries 0.205}\\
\end{tabular}
\end{minipage}
}
\hfill
\subfloat[
Query embedding lengths.
\label{tab:abla_N}
]{
\begin{minipage}[t]{0.31\linewidth}
\centering
\tablestyle{2pt}{1.1}
\scriptsize
\begin{tabular}{c  cc c| c}
    \multicolumn{1}{c|}{\multirow{2}*{\makecell{Query \\ length $N$}}}  &\multicolumn{3}{c|}{Video-CraftBench} &CALVIN  \\
      \multicolumn{1}{c|}{} & Paper &Block & LPIPS$\downarrow$ &Avg.Len.  \\
    \shline
    \specialrule{0em}{0pt}{1pt} 
    \multicolumn{1}{c|}{baseline} &0.0 &28.5 &0.312 &1.11\\
    \multicolumn{1}{c|}{1} &41.9 &65.0  &0.221  &1.53\\
    \multicolumn{1}{c|}{2} &55.1 &69.5  &0.210 &1.64 \\
    \multicolumn{1}{c|}{4} &{\bfseries 68.8} &{\bfseries 77.5} &\underline{0.205} &\underline{1.87}  \\
    \multicolumn{1}{c|}{8} &\underline{65.0} &{\bfseries 77.5}  &{\bfseries 0.195} &{\bfseries 1.88}\\
\end{tabular}
\end{minipage}
}
\hfill
\subfloat[
LDM/VDM interplay.
\label{tab:abla_interplay}
]{
\centering
\begin{minipage}[t]{0.26\linewidth}
\centering
\tablestyle{1pt}{1.1}
\scriptsize
\begin{tabular}{llll}
Proj &Interplay &\multirow{2}*{Paper} &\multirow{2}*{Block} \\
Layer &Attn type & & \\
\shline
\specialrule{0em}{0pt}{1pt}
MLP  &cross &52.0 &61.3 \\
+self &cross &52.3 &61.8\\
MLP  &causal cross &69.8 &78.6 \\
+self &causal cross &{\bfseries 72.3} &{\bfseries 80.9}\\
& & & \\
\end{tabular}
\end{minipage}
}
\hfill
\\
\vspace{0.9em}
\hfill
\subfloat[
Codebook sizes of dLDM.
\label{tab:abla_codebooksize}
]{
\begin{minipage}[t]{0.35\linewidth}
\centering
\tablestyle{3pt}{1.1}
\scriptsize
 \begin{tabular}{ c | ccc | c}
                  \multirow{2}*{\makecell{Codebook \\ size  }}  \multirow{2}*  & \multicolumn{3}{c|}{Video-CraftBench} &\multicolumn{1}{c}{CALVIN} \\
                 
                     &Paper &Block  &LPIPS$\downarrow$ &Avg.Len.  \\
                \shline
                \specialrule{0em}{0pt}{1pt}
                 baseline &0.0 &28.5 &0.312 &1.11 \\
                  8   &20.1  &29.9 &0.258  &1.65 \\
                   1000  &{\bfseries 68.8} &{\bfseries 77.5}  &{\bfseries 0.205} &1.87 \\
                  4096   &\underline{50.4} &\underline{59.6} &\underline{0.208}  &{\bfseries 1.90}\\
                  64,000  &29.4 &36.0 &0.230 &\underline{1.89} \\
            \end{tabular}
\end{minipage}
}
\hfill
\subfloat[
Compression lengths of dLDM.
\label{tab:abla_T}
]{
\begin{minipage}[t]{0.32\linewidth}
\centering
\tablestyle{3pt}{1.1}
\scriptsize
\begin{tabular}{c | cc| c}
     \multirow{2}*{\makecell{Length \\  $T$}} &\multicolumn{2}{c|}{Video-CraftBench} &CALVIN \\
     &Paper &Block   &Avg.Len.  \\
    \shline
    \specialrule{0em}{0pt}{1pt}
    2   &19.1 &38.7   &1.55  \\
    9   &55.4 &68.7   &1.61 \\

    49   &65.3 &76.2  &\underline{1.80} \\
    
    93   &\underline{68.8} &{\bfseries 77.5}   &{\bfseries 1.87}  \\
    177  &{\bfseries 69.0} &\underline{76.8}  &1.79   
\end{tabular}
\end{minipage}
}
\hfill
\subfloat[
Training strategies of VDM.
\label{tab:abla_vdm}
]{
\begin{minipage}[t]{0.26\linewidth}
\centering
\tablestyle{3pt}{1.1}
\scriptsize
\begin{tabular}{c  cc }
     \multicolumn{1}{c|}{\multirow{2}*{\makecell{Training \\ Strategy}}} &\multicolumn{2}{c}{Video-CraftBench}  \\
     \multicolumn{1}{c|}{} &Paper &Block    \\
    \shline
    \specialrule{0em}{0pt}{1pt}
    \multicolumn{1}{c|}{baseline}   &0.0 &28.5    \\
    \multicolumn{1}{c|}{random}   &0.0 &0.0    \\
    \multicolumn{1}{c|}{freeze}   &31.7 &40.2  \\

    \multicolumn{1}{c|}{lora}   &50.9 &62.3\\
    
    \multicolumn{1}{c|}{full}   &68.8 &77.5   \\
    
\end{tabular}
\end{minipage}
}
\hfill
\caption{\textbf{Ablation studies.} We train with Video-CraftBench only and evaluate CALVIN in its in-domain setting.}
\vspace{-0.5em}
\label{tab:ablations} 
\end{table*}


    
\begin{figure}[t]
\includegraphics[width=0.95\linewidth]{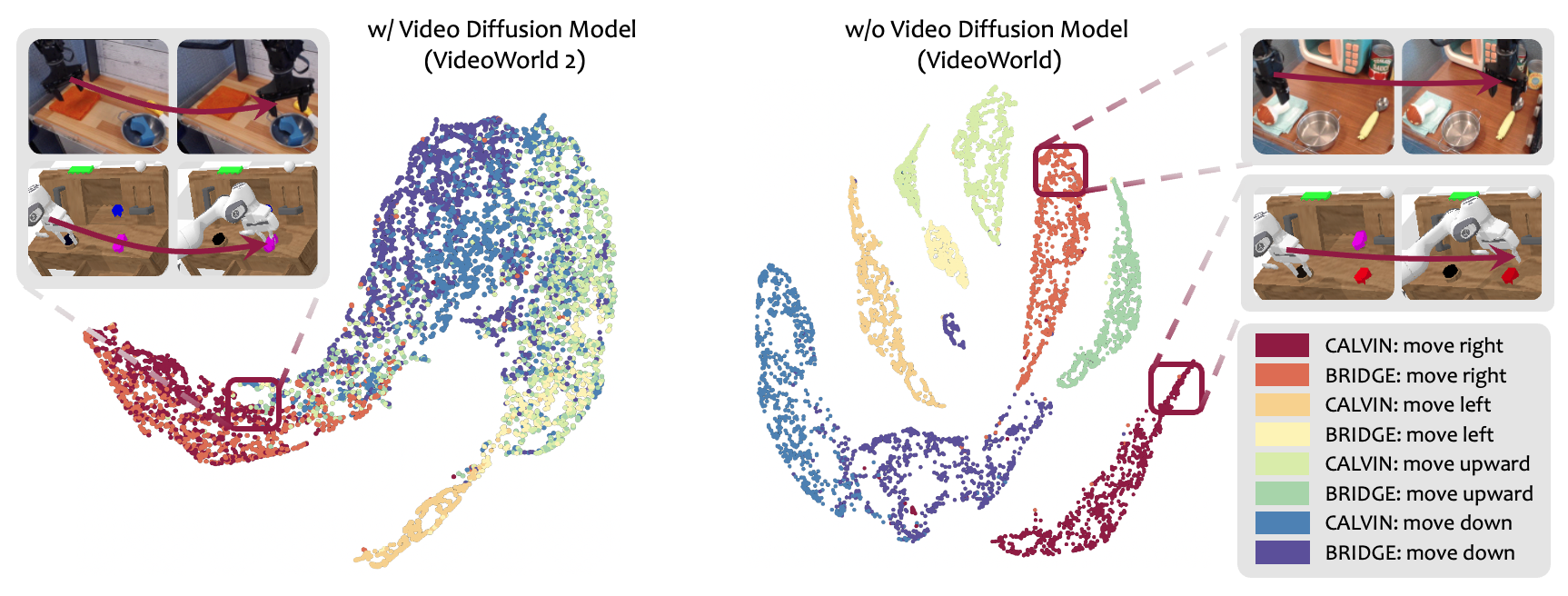}
\centering
\caption{\textbf{UMAP of latent codes.} 
With a pretrained VDM, latent codes for the same action align more tightly than those in VideoWorld~\cite{ren2025videoworld} across environments, indicating more consistent and transferable dynamics.  (Left) In distinct environments like Bridge~\cite{walke2023bridgedata} and CALVIN~\cite{mees2022calvin}, latent representations of the same action (e.g., the robotic arm moving right) are highly similar. (Right) Conversely, in VideoWorld, latent codes for the same action exhibit significant divergence across environments and fail to cluster effectively in the UMAP visualization.
}
\label{fig:umap}
\vspace{-1em}
\end{figure}

\begin{figure*}[t]
\includegraphics[width=\linewidth]{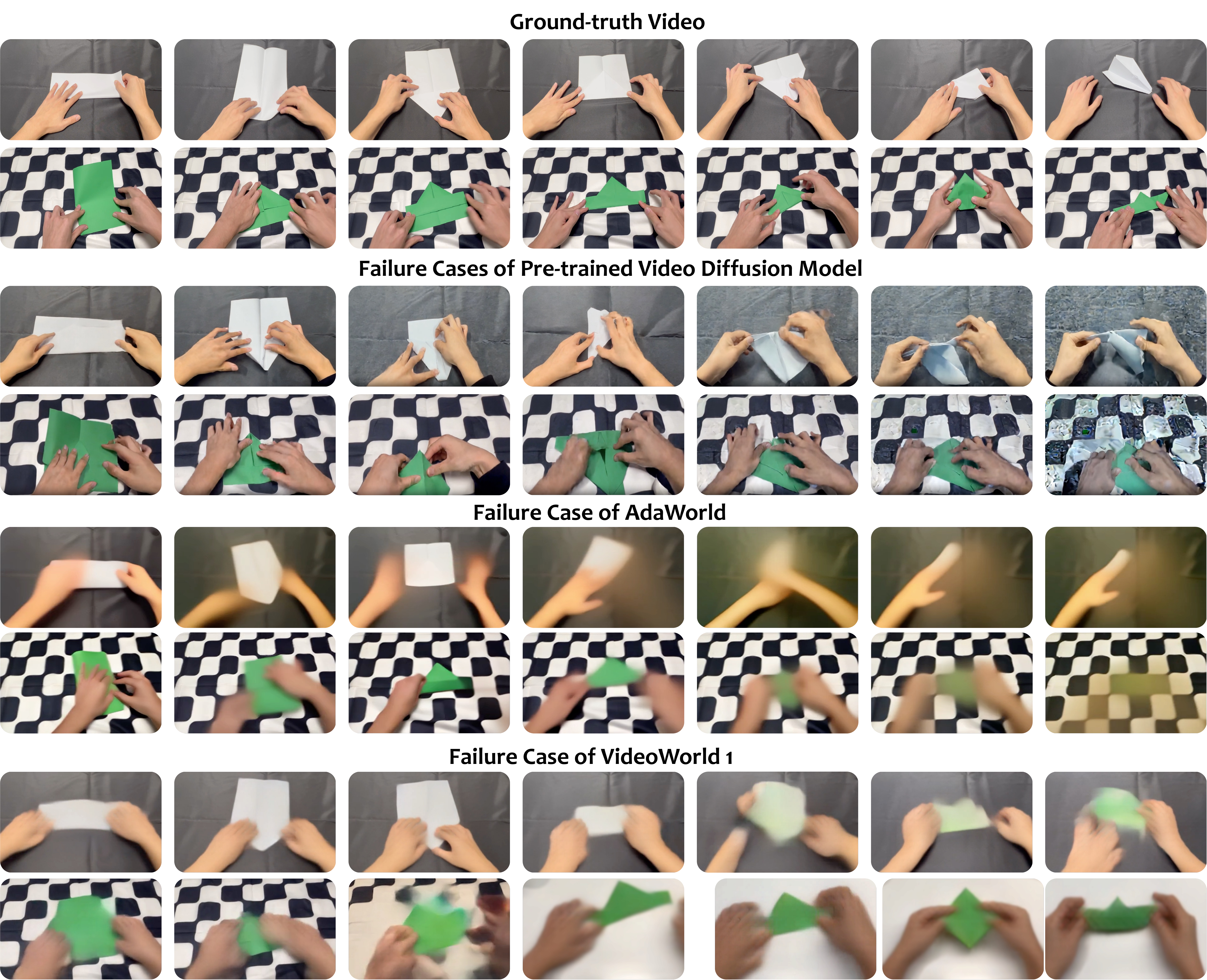}
\centering
\caption{\textbf{Visualization of failure cases } in other baselines on the Video-CraftBench.   Best viewed in color.
}
\label{fig:supp_fail_vis}
\vspace{-1em}
\end{figure*}

\noindent{\textbf{Query embedding length of dLDM.}} Tab.~\ref{tab:abla_N} ablates the number of dLDM query embeddings $N$. A larger $N$ captures more information but also risks encoding noise and increases the sequence length for the AR transformer. We find that $N=1$ already achieves a respectable task success rate, while $N=4$ provides the best performance balance. Increasing $N$ to 8 offers a slight LPIPS benefit but introduces more noise, which reduces the overall success rate. 

\noindent{\textbf{Interplay between VDM and LDM.}} Tab.~\ref{tab:abla_interplay} ablates the interplay mechanism between the LDM and VDM, which consists of a projection layer (MLP and causal self-attention) and causal cross-attention. The causal cross-attention ensures generation relies solely on the current time step's latents, effectively preventing information leakage. Results show that incorporating self-attention into the projection layer and utilizing causal cross-attention significantly enhance LDM-VDM interaction, validating the effectiveness of our design

\noindent{\textbf{Horizon length in dLDM.}} Tab.~\ref{tab:abla_T} ablates the dLDM context length $T$.  At $T=2$ (a LAPA-like setting), codes remain transferable, but performance on the long-range paper folding task is poor due to a lack of temporal perception. This minimal context also provides low-quality motion guidance for the VDM, leading to poor LPIPS scores. Performance improves as $T$ increases, plateauing at $T=93$, which corresponds to the maximum context length of our Cosmos VDM.

\noindent{\textbf{dLDM codebook size.}} Tab.~\ref{tab:abla_codebooksize} examines the impact of the dLDM codebook size (adjusted via FSQ levels) on policy learning.  We find that CALVIN, with its simpler action space, benefits from a relatively small codebook. In contrast, the more complex Video-CraftBench requires a larger codebook to achieve significant improvements. However, an excessively large codebook risks encoding extraneous noise and can hinder dLDM training convergence, a finding consistent with VideoWorld~\cite{ren2025videoworld}.

\noindent{\textbf{Training strategies of VDM.}} In Tab.~\ref{tab:abla_vdm}, we examine different training configurations for the video diffusion model. Our default approach fully fine-tunes the pre-trained VDM to exploit its appearance priors. In contrast, training a randomly initialized VDM (row 1) leads to model collapse and fails to generate valid videos. We also experiment with freezing the pre-trained VDM and updating only the VQ-VAE components (encoder, quantizer, and projection layer). Although this setup yields performance gains, it falls significantly short of LoRA or full fine-tuning. We believe this gap exists because the VDM needs further adaptation to capture the fine-grained manipulation details specific to Video-CraftBench.

\section{Further Discussion with Other Works}
We provide a comprehensive analysis, highlighting the differences between VideoWorld 2 and prior works in both objectives and methodology.

\textbf{Regard task objectives.} Most prior works focus on short-horizon prediction, simulated control, or reconstruction-centric tasks. In contrast, VideoWorld 2 targets complex, minute-long real-world tasks, where appearance variation and long-horizon error accumulation are dominant. This exposes appearance interference as a critical bottleneck, necessitating our dLDM.

\textbf{Regarding ``disentanglement''.} Prior works~\cite{wang2025vidtwinvideovaedecoupled, wu2023pretrainingcontextualizedworldmodels, wu2024ivideogpt, chen2024igor, ren2025videoworld, lapa_ye2024latent} rely on VAE-style reconstruction objectives to separate dynamics from appearance. As demonstrated in VideoWorld~2, this approach is insufficient for complex real-world videos, as reconstruction objectives compel latent codes to encode task-irrelevant details. VideoWorld~2 instead mitigates appearance interference by leveraging a pretrained VDM for appearance modeling, forcing latents to focus on task-relevant dynamics. While CoLA~\cite{wang2025coevolvinglatentactionworld} also employs a VDM, it is limited to short 2-frame transitions. In contrast, VideoWorld~2 models multi-step dynamics and reuses coarse VAE decoder outputs to provide structured temporal cues—a mechanism critical for long-horizon tasks (Tab.~\ref{tab:abla_arch}, rows 3 vs. 5). Replacing our dLDM with prior works' tokenizers~\cite{wu2024ivideogpt} or adopting CoLA's training scheme significantly degrades long-sequence performance, validating our design.

\textbf{Regarding the effect of the VDM.} The role of the VDM in VideoWorld~2 also differs from prior approaches. Works such as $\epsilon$-VAE~\cite{zhao2025epsilonvae} and DiVoT~\cite{ge2024divot} utilize VDMs for pixel-level reconstruction, necessitating that latent codes encode both appearance and dynamics. Others, like IGOR~\cite{chen2024igor} and AdaWorld~\cite{adaworld}, employ VDMs primarily for rendering. This disconnects the VDM from latent optimization, failing to improve task objectives. In contrast, VideoWorld~2 offloads appearance modeling to the VDM, compelling latents to focus exclusively on task-relevant dynamics. Compared to CoLA~\cite{wang2025coevolvinglatentactionworld}, VideoWorld~2 benefits from multi-frame modeling and a VDM architecture specifically tailored for long-horizon learning.

\section{Conclusion}
\label{sec:conclusion}
In this work, we explore knowledge learning for complex, long-horizon tasks from raw videos through experiments on Video-CraftBench and in robotic manipulation environments.  We find that decoupling visual appearance from core actions is crucial for real-world knowledge learning. Based on this finding, we propose VideoWorld~2, a model featuring a dynamics-enhanced latent dynamics model (dLDM) that leverages a pre-trained VDM to learn generalizable, transferable policies directly from video. While our method shows significant potential, we leave its continued scaling to future work, moving toward the goal of enabling AI to learn the vast knowledge encapsulated in the real world.

\clearpage

\bibliographystyle{plainnat}
\bibliography{main}

\clearpage

\beginappendix



In this supplementary material, we provide additional details and analyses:
\begin{itemize}
    \item Sec.~\ref{sec:supp_detail} describes the training setup, the structure of the dLDM, and other implementation details.
    \item Sec.~\ref{sec:supp_data} provides further analysis of Video-CraftBench.
    \item Sec.~\ref{sec:supp_vis} provides additional visualizations of both our method and the baselines.
\end{itemize}

\section{Implementation Details}
\label{sec:supp_detail}

\noindent \textbf{Training details.} We present the detailed training configurations of dynamics-enhanced latent dynamics model and auto-regressive transformer in Tab.~\ref{table:supp_training_setting}.

\noindent \textbf{Dynamics-enhanced latent dynamics model (dLDM).}
As shown in Fig.~4, the proposed dLDM consists of four components. 
(i) A causal encoder extracts visual features. 
(ii) A set of learnable queries attends to the encoder features to extract visual changes and produce the latent dynamic codes. 
(iii) A decoder reconstructs frames from the latent dynamic codes and the feature of the first frame, providing coarse motion cues. 
(iv) A pretrained Video Diffusion Model (VDM) generates high-fidelity future frames by taking three inputs: the first frame, the low-resolution decoder outputs, and the latent dynamic codes. The latent codes are injected into the VDM via causal cross-attention, allowing the codes to concentrate on task-relevant dynamics while the VDM handles visual appearance modeling. We provide PyTorch-style pseudocode for the full dLDM and its components in Alg.~\ref{alg:code}.

\noindent \textbf{Latent pre-training on CALVIN.}
In Sec.~5.4 (line 432), we present a latent pre-training experiment where latent dynamic tokens serve as prediction targets. This enables the model to learn manipulation knowledge from unlabeled data, thereby facilitating better adaptation to ground-truth actions. Following LAPA~\cite{lapa_ye2024latent}'s protocol, given a task trajectory $x_{0:T}$, we train the AR Transformer to predict the quantized latent dynamic embeddings $\{z_k^n\}^{K,N}_{k=1,n=1}$ (defined in Sec.~3 line 198), conditioned on the language instruction and the first image $x_0$. We extend the AR transformer's vocabulary to include these latent dynamic tokens. Since this pre-training does not rely on ground-truth actions, it can use \textit{any unlabeled video data.}

Following latent pre-training, we further fine-tune the model with ground-truth labels to derive actions executable in the simulation environment. We append a new action head (an MLP layer) to the transformer to map the hidden states, which are originally used for latent token prediction, to real actions via an $\ell_2$ loss. To ensure compatibility with cross-environment data, we exclusively use CALVIN's static camera images during both pre-training and fine-tuning, omitting wrist images and state parameters.  This latent pre-training should provide the AR transformer with a manipulation prior, leading to better fine-tuning performance.

\section{Details in Video-CraftBench}
\label{sec:supp_data}

\noindent \textbf{Key steps for evaluation.} In Sec.~4.2 (line 314), we define seven key steps in the paper folding task to evaluate the task success rate. Fig.~\ref{fig:supp_key_step} (top) provides a schematic diagram of these key steps.


\noindent \textbf{Train and test environments.}  Fig.~\ref{fig:supp_key_step} (bottom) illustrates the Video-CraftBench training and testing environments. To evaluate the model's generalization to new environments, the test set features diverse variations compared to the training set. Specifically, the paper folding test set differs in background, texture, paper appearance, and camera viewpoints, while the block-building varies in initial block arrangement, color combinations, and camera angles.

\begin{table}[t]
    \footnotesize
    \centering
    {
\begin{tabular}{l|l |l}
     Config & dLDM & AR Transformer\\

    \shline
    optimizer & AdamW &AdamW\\
     base learning rate & 1e-4 &3e-4\\
     weight decay & 0.1  &0.05\\
     optimizer monmentum & $\beta_1$, $\beta_2$=0.9,0.99 & $\beta_1$, $\beta_2$=0.9,0.98 \\
     batch size &128 & 256 \\
     learning rate schedule & WarmipDecayLR & WarmipDecayLR \\
     warmup iterations & 2e+3 &5e+3\\
     max iterations &1e+5 &5e+4 \\
     augmentations & None & None \\
     Training Loss & L2 \& Denoising loss &CE loss  \\
     Training Target & Recon. \& Denoising &Next token pred.  \\

\end{tabular}
}
\caption{\textbf{Training configurations} for the dynamics-enhanced latent dynamics model and auto-regressive transformer.}
\vspace{2em}
\label{table:supp_training_setting}
\end{table}

\noindent \textbf{Task analysis.} Fig.~\ref{fig:data_time} shows the task duration distribution of Video-CraftBench. The largest portion (37.3\%) falls within 45–60 seconds, followed by 27.1\% at 60–90 seconds, while short tasks (20–30s) make up only 10.9\%. This distribution highlights our emphasis on long-horizon tasks. Regarding task types (Fig.~\ref{fig:data_task}), paper folding accounts for 55.3\% (5.2 hours) of the total duration due to its longer execution time, while block building comprises 1.8 hours.

\noindent \textbf{Task classifier of Video-CraftBench.} We detail the training of the classifier used to evaluate key step completion in Sec.~4.2 (line 314). Its primary objective is to determine whether the model has reached key steps by analyzing paper shapes and block arrangements. Appearance distractions, such as texture, color, and the background drift shown in Fig.~\ref{fig:supp_fail_vis} (last row), are disregarded. These visual quality is assessed separately via LPIPS and SSIM. Therefore, we initialize the classifier with DINOv2-Base (86M parameters), chosen for its exceptional geometric awareness and ability to extract fine-grained features. 

To train this classifier, we construct a dataset of $\sim$25k labeled frames. First, we extract $\sim$15k frames of both key and non-key steps from the training and testing environments. To further enhance diversity and mitigate bias, we generate task execution videos using all models involved in the evaluation, manually verify successful trajectories, and extract an additional $\sim$10k frames.  We then attach a classification head to the DINOv2 backbone and fine-tune the entire model. The classifier achieves a test accuracy of 96.1\%, ensuring reliable judgment for subsequent evaluations.


\begin{figure}[t]
\begin{minipage}{0.5\linewidth}
\centering
\includegraphics[width=0.9\linewidth]{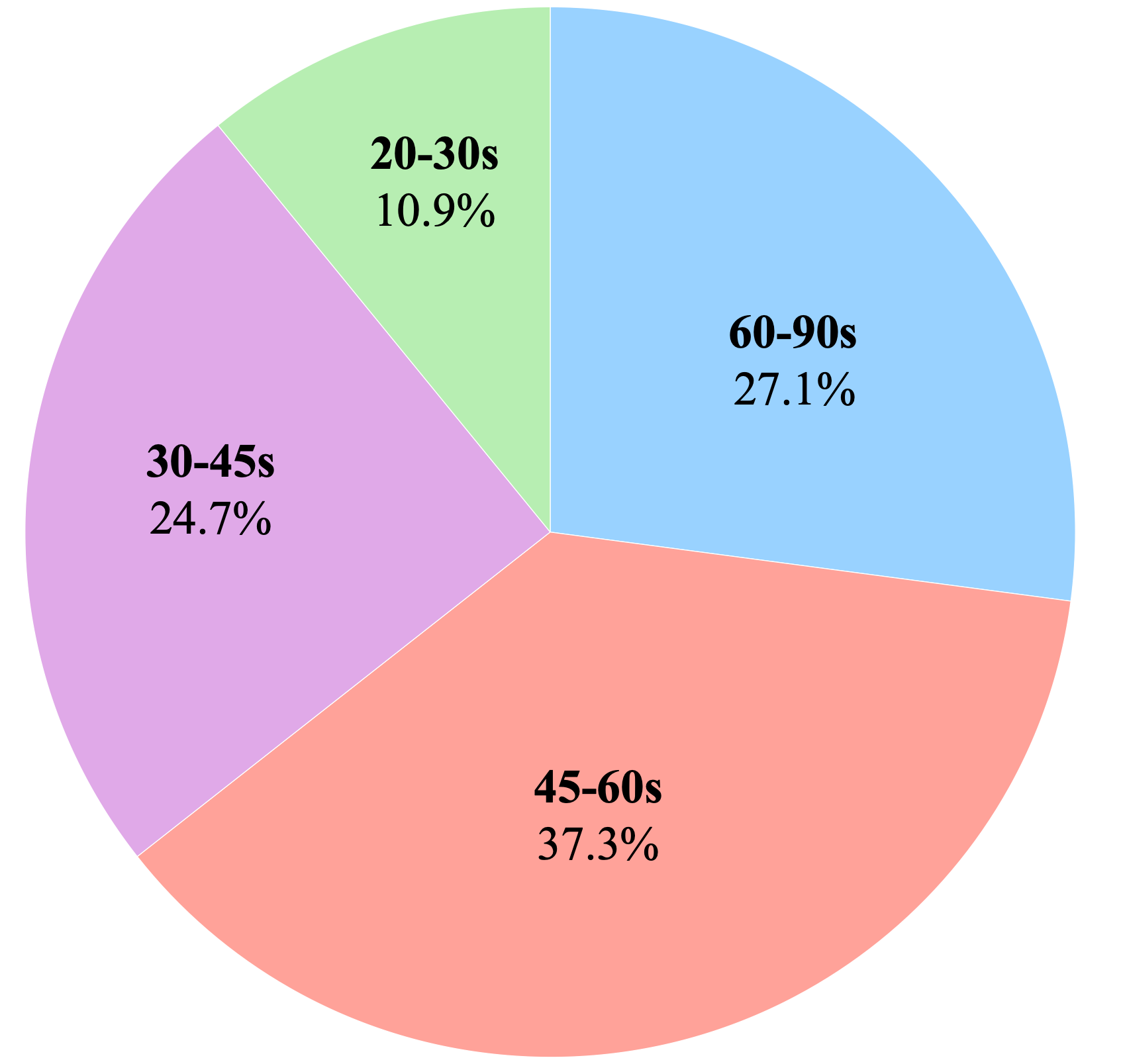} 
\caption{\textbf{Video duration distribution}.}
\label{fig:data_time}
\end{minipage}
\begin{minipage}{0.5\linewidth}
\centering
\includegraphics[width=0.9\linewidth]{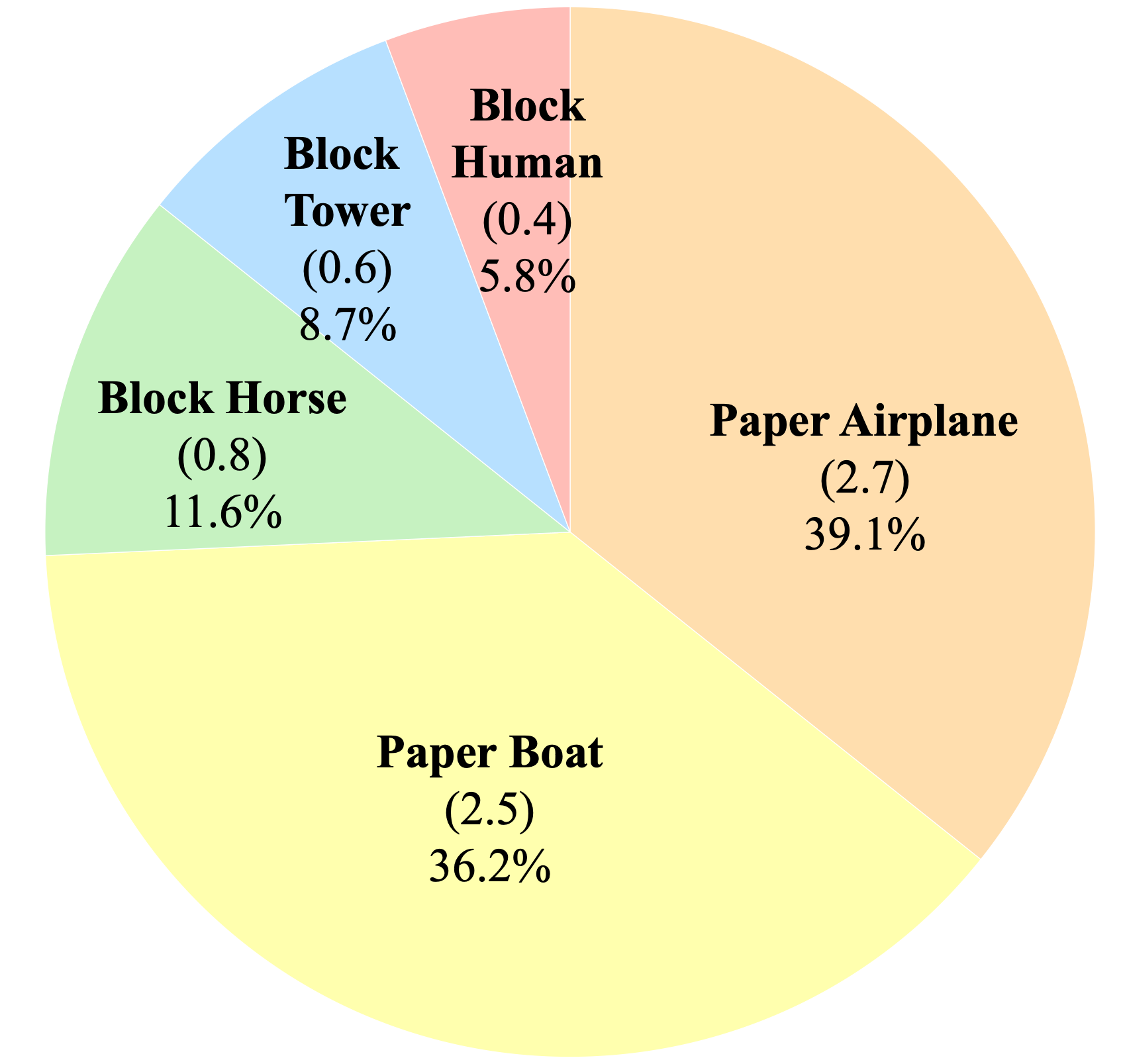} 
\caption{\textbf{Task type distribution}.}
\label{fig:data_task}
\end{minipage}
\hfill
\vspace{2em}
\end{figure}

\SetKwSty{textbf} 
\SetFuncSty{textbf}
\SetArgSty{textnormal}
\SetFuncSty{textnormal}

\begin{algorithm2e}[H]
\caption{dLDM Algorithm (No Underline Hack)}
\label{alg:dldm_hack}
\SetAlgoLined
\DontPrintSemicolon

\SetKwProg{Fn}{Function}{}{end} 

\SetKwSty{textbf}
\SetArgSty{textnormal}

\SetKwInOut{Input}{Input}
\SetKwInOut{Variables}{Variables}

\Input{Video sequence $V \in \mathbb{R}^{(1+T) \times h \times w \times 3}$}
\Variables{Learnable embeddings $Q_{ldm} \in \mathbb{R}^{N \times C}$}

\vspace{0.3em}


\Fn{}{ \textbf{encoder}(video): \tcp*[f]{Encoder Function}
    \;
    $f \leftarrow \text{Causal3DCNN}(video)$ \;
    \For{layer \textbf{in} encoder\_layers}{
        $f \leftarrow layer(f)$ \;
    }
    \Return $z, f[0]$
}

\vspace{0.3em}

\Fn{}{ \textbf{ldm\_qformer}(f):
    \;
    $q\_list \leftarrow []$ \;
    \For{$k \leftarrow 2$ \KwTo $K$}{
        $f_k \leftarrow f[:k]$ \;
        $q_k \leftarrow \text{MLP}(\text{CrossAttention}(Q_{ldm}, f_k, f_k))$ \;
        $q\_list.\text{append}(q_k)$ \;
    }
    \Return $\text{stack}(q\_list)$
}

\vspace{0.3em}

\Fn{}{ \textbf{decoder}(z, first\_f):
    \;
    $rec \leftarrow \text{repeat}(first\_f, K, \text{dim}=0)$ \;
    \For{\_ \textbf{in} decoder\_layers}{
        $rec \leftarrow \text{Causal3DCNN}(rec)$ \;
        $rec \leftarrow \text{up\_scale}(rec)$
    }
    \Return $rec$
}

\vspace{0.3em}

\Fn{}{ \textbf{main}(video):
    \;
    $z, first\_f \leftarrow \text{encoder}(video)$ \;
    $z \leftarrow \text{FSQ}(z)$ \;
    \If{is\_train}{
        $rec \leftarrow \text{decoder}(z.\text{detach}(), first\_f)$ \;
        \Return $\text{MSE}(rec, video) + \text{VDM}(video, z, rec)$
    }
    \Return $z$
}

\end{algorithm2e}

\begin{figure*}[t]
\includegraphics[width=\linewidth]{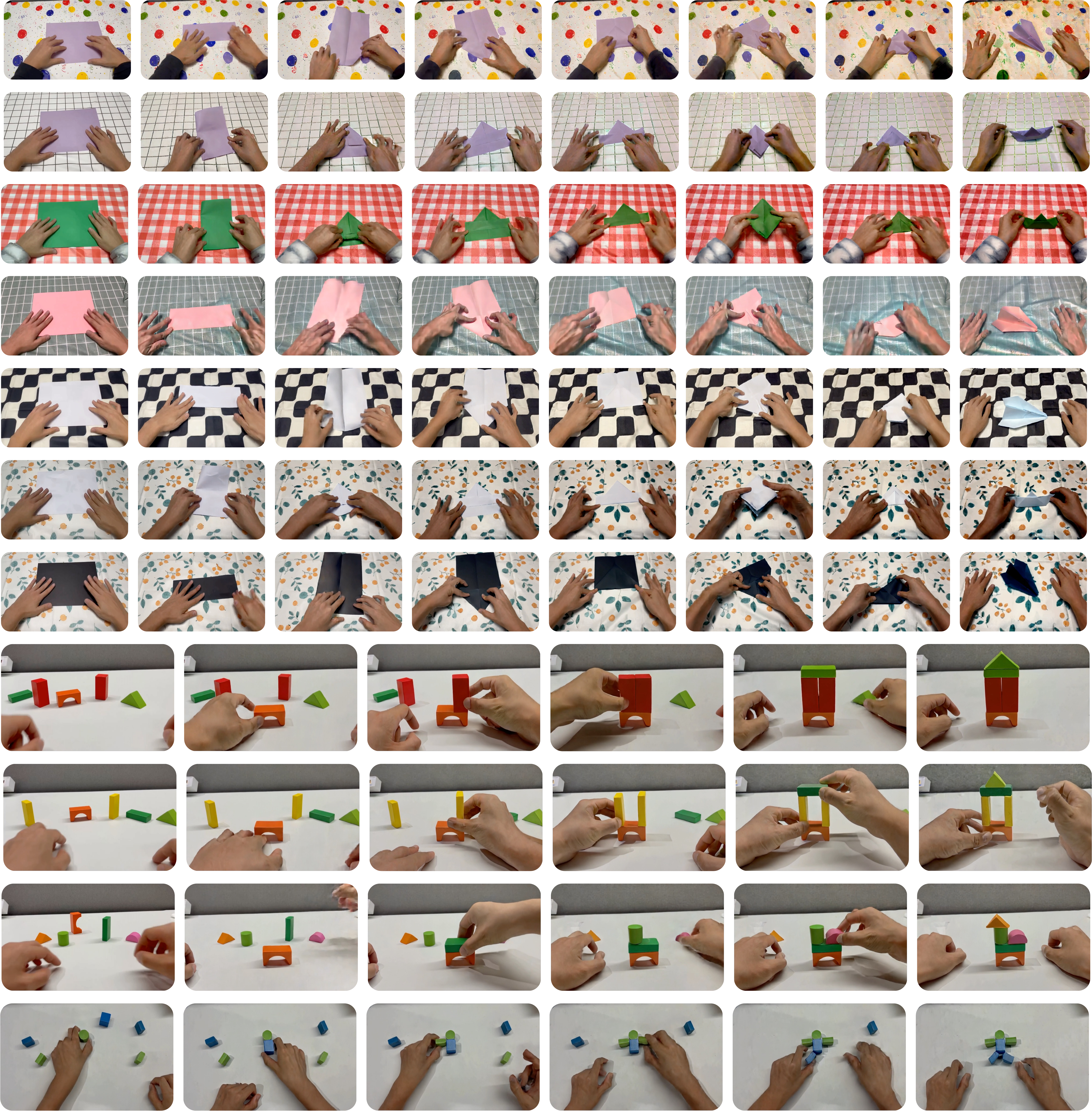}
\centering
\caption{\textbf{More visualization of VideoWorld~2 on Video-CraftBench.}
}
\label{fig:supp_more_vis}
\vspace{-4mm}
\end{figure*}

\section{Visualizations}
\label{sec:supp_vis}

Fig.~\ref{fig:supp_more_vis} presents additional visualizations of VideoWorld~2 on Video-CraftBench. We also include more video demonstrations in the supplementary material. Thanks to the dLDM, VideoWorld~2 generates accurate long-horizon execution sequences in novel environments, producing visually coherent, high-quality videos. Note that the VDM processes only 93 frames at a time, while full task sequences can span thousands of frames. Long videos are therefore generated auto-regressively by extending each segment from the final frame of the previous one. Because the VDM’s inherent reconstruction noise accumulates over time, visual artifacts such as lighting, texture, or color shifts may gradually appear. Nevertheless, the key steps within the generated sequences remain accurate.

\end{document}